\documentclass[nohyperref]{article}

\usepackage{microtype}
\usepackage{graphicx}
\usepackage[center]{caption}
\usepackage{caption}
\usepackage{subcaption}
\usepackage{bm}
\usepackage{booktabs} %
\usepackage{subfiles}
\usepackage{xcolor}

\usepackage{enumitem}

\usepackage{hyperref}

\usepackage[accepted]{icml2023}

\newenvironment{new_content}{\par\color{black}}{\par}
\newcommand{\upd}[1]{\textcolor{black}{#1}}

\usepackage{amsmath}
\usepackage{amssymb}
\usepackage{mathtools}
\usepackage{amsthm}
\usepackage{multirow}

\usepackage{graphicx}

\newcommand{\x}{{\bm{x}}}
\newcommand{\y}{{\bm{y}}}
\renewcommand{\d}{{\rm{d}}}

\newcommand{\E}{\mathbb{E}}
\newcommand{\eps}{\bm{\epsilon}}
\newcommand{\method}{\textsc{ReDi}}

\usepackage[capitalize,noabbrev]{cleveref}

\theoremstyle{plain}
\newtheorem{theorem}{Theorem}[section]

\theoremstyle{definition}
\newtheorem{definition}[theorem]{Definition}
\newtheorem{assumption}[theorem]{Assumption}
\theoremstyle{remark}

\usepackage[textsize=tiny]{todonotes}

\icmltitlerunning{ReDi: Efficient Learning-Free Diffusion Inference via Trajectory Retrieval\hfill\thepage}

\begin{document}

\twocolumn[
\icmltitle{ReDi: Efficient Learning-Free Diffusion Inference via Trajectory Retrieval}

\icmlsetsymbol{equal}{*}

\begin{icmlauthorlist}
\icmlauthor{Kexun Zhang}{yyy}
\icmlauthor{Xianjun Yang}{yyy}
\icmlauthor{William Yang Wang}{yyy}
\icmlauthor{Lei Li}{yyy}
\end{icmlauthorlist}

\icmlaffiliation{yyy}{Department of Computer Science, University of California, Santa Barbara}

\icmlcorrespondingauthor{Kexun Zhang}{kexun@ucsb.edu}

\icmlkeywords{Machine Learning, ICML}

\vskip 0.3in
]

\printAffiliationsAndNotice{}  %

\begin{abstract}
Diffusion models show promising generation capability for a variety of data. Despite their high generation quality, the inference for diffusion models is still time-consuming due to the numerous sampling iterations required. To accelerate the inference, we propose \textsc{ReDi}, a simple yet learning-free \textbf{\textit{Re}}trieval-based \textbf{\textit{Di}}ffusion sampling framework. From a precomputed knowledge base, \textsc{ReDi} retrieves a trajectory similar to the partially generated trajectory at an early stage of generation, skips a large portion of intermediate steps, and continues sampling from a later step in the retrieved trajectory. We theoretically prove that the generation performance of \textsc{ReDi} is guaranteed. Our experiments demonstrate that \textsc{ReDi} improves the model inference efficiency by 2$\times$ speedup. Furthermore, \textsc{ReDi} is able to generalize well in zero-shot cross-domain image genreation such as image stylization. \upd{The code and demo for \method~is available at \url{https://github.com/zkx06111/ReDiffusion}.}
\end{abstract}

\section{Introduction}

Deep generative models are changing the way people create content. Among them, diffusion models have shown great capability in a variety of applications including image synthesis \cite{ho2020denoising, dhariwal2021diffusion}, speech synthesis \cite{liu2022diffsinger}, and point cloud generation \cite{zhou20213d}. Latent diffusion models \cite{rombach2022high} such as Stable Diffusion are able to generate high-quality images given text prompts. However, the basic sampler for diffusion models proposed by \citet{ho2020denoising} requires a large number of function estimations (NFEs) during inference, making the generation process rather slow. For example, the basic sampler takes 336 seconds on average to run on an NVIDIA 1080Ti, where improving the efficiency of diffusion model inference is crucial.

Previous studies on improving the efficiency of diffusion model inference can be categorized into two types, learning-based ones, and learning-free ones. Learning-based samplers \cite{salimans2021progressive, meng2022distillation, lam2022bddm, watson2021learning, kim2022dmcmc} require additional training to reduce the number of sampling steps. However, their training is expensive, especially for large diffusion models like Stable Diffusion. In contrast, learning-free samplers do not require training, and are, therefore, applicable to more scenarios. In this paper, we focus on learning-free approaches to speed up inference.

Existing efficient learning-free samplers for diffusion \cite{liu2021pseudo, lu2022dpm, lu2022dpms, zhang2022fast, karras2022elucidating} all try to find a more accurate numerical solver for the diffusion ODE \cite{song2021score}, but they do not utilize its sensitivity. The sensitivity of ODEs suggests that under certain conditions, a small perturbation in the initial value does not change the solution too much \cite{khalil2002nonlinear}. This observation motivates the assumption that a previously generated trajectory - if close enough to the current trajectory - can serve as an estimate for it.

\begin{figure}[tpb!]\vskip 0.1in
  \centering
  \includegraphics[width=0.45\textwidth]{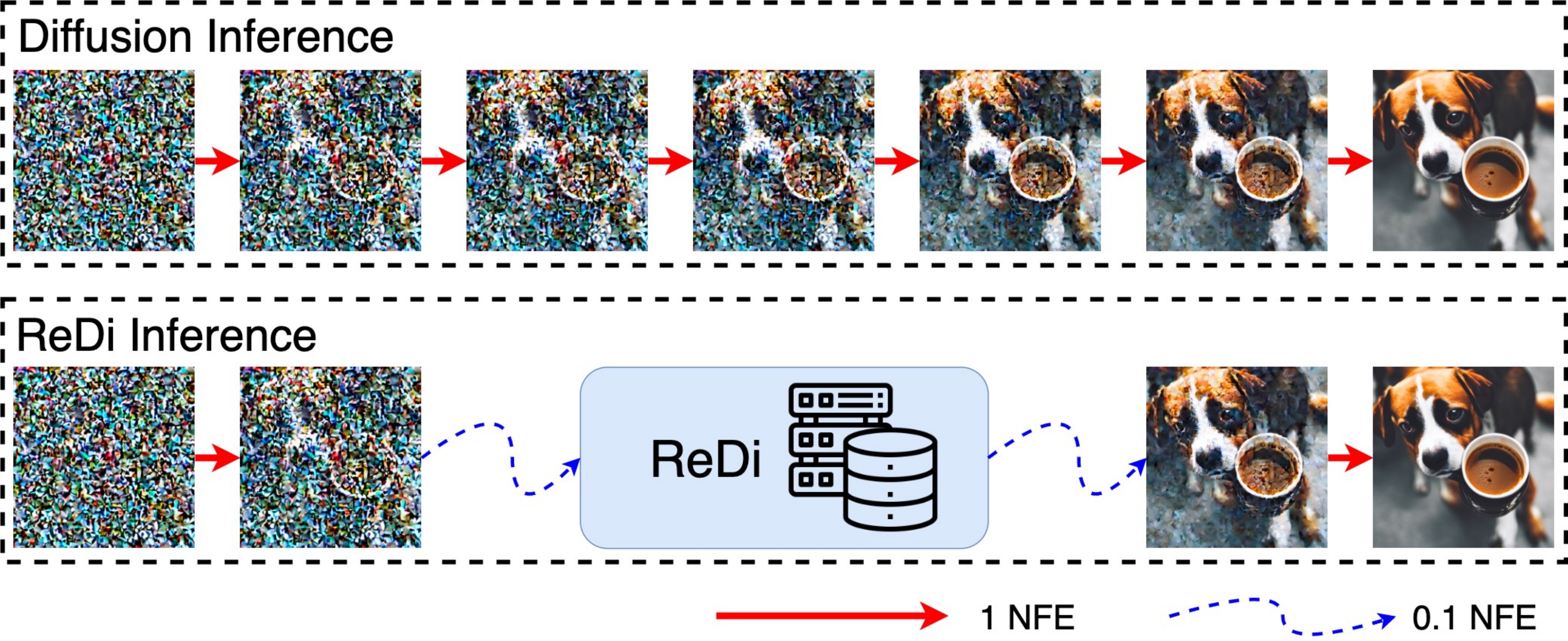}
  \caption{Diffusion Inference (upper) and \textsc{ReDi} Inference (lower). \textsc{ReDi} reduces the number of function estimations (NFEs) during inference by skipping several intermediate sampling steps. The overhead brought by \textsc{ReDi} is minimal compared to the cost it reduced.}\label{fig:cat_dog}\vskip 0.1in
\end{figure}

In this paper, we propose \textsc{ReDi}, a learning-free sampling framework based on trajectory retrieval. \autoref{fig:cat_dog} illustrates the original full diffusion inference and the \textsc{ReDi} inference. Given a pre-trained diffusion model, \textsc{ReDi} does not modify its weights or train new modules. Instead, we fix the model and build a knowledge base $\mathcal{B}$ of trajectories upon a chosen dataset during the precomputation. During the inference, \textsc{ReDi} first computes the early few steps in the trajectory as they are, and then retrieves a similar trajectory from $\mathcal{B}$. In this way, one later step in the retrieved trajectory can surrogate the actual one and serve as an initialization point for the model. By jumping from an early time step to a later time step $V$, \textsc{ReDi} is able to save a larger portion of function estimations (NFEs)  any numerical solver.

We first prove that \textsc{ReDi}'s performance is bounded by the distance between the query trajectory and the retrieved neighbor. Then we report results from in-domain experiments to show empirically that with a moderate-sized knowledge base, \textsc{ReDi} is able to achieve comparable performance to recent efficient samplers with a $2\times$ speedup. To demonstrate that \textsc{ReDi} generalizes well in cross-domain adaptation, we propose an extension to \textsc{ReDi} that generates various stylistic images given the same single-style knowledge base. The stylized images generated by \textsc{ReDi} are well-rated by human evaluators. Our ablation studies show that under different settings, the actual results of \textsc{ReDi} correlate well with the theoretical bounds, indicating the bounds are tight enough to estimate the generation quality.

Our contributions are as follows:
\begin{itemize}[noitemsep, leftmargin=*, topsep=2pt]
    \item We propose \textsc{ReDi}, a retrieval-based learning-free framework for diffusion models. \textsc{ReDi} is orthogonal to previous learning-free samplers and reduces the number of function estimations (NFEs) by skipping some intermediate steps in the trajectory.
    \item We prove a theoretical bound for the generation quality of \textsc{ReDi} that correlates well with the actual performance.
    \item We show empirically that \textsc{ReDi} can improve the inference efficiency with precomputation and perform well in zero-shot domain adaptation.
\end{itemize}

\section{Related Work}

\noindent \textbf{Retrieval-Based Diffusion Models~}
Previous studies on retrieval-based diffusion~\cite{blattmann2022semi, sheynin2022knn} have different emphases including rare entity generation \cite{chen2022re}, out-of-domain adaptation, semantic manipulation, and parameter efficiency.
However, they all need to train a new model instead of building upon a trained model, which requires much computing power and time. They retrieve images and/or text using pre-built similarity measures like CLIP embedding cosine similarity~\cite{radford2021learning}. But the pre-built measures they use are not specially designed for diffusion models and have no proven performance guarantee.

\noindent \textbf{Learning-Free Diffusion Samplers~}
Most learning-free diffusion samplers are based on the stochastic/ordinary differential equation (SDE/ODE) formulation of the denoising process proposed by \citet{song2021score}. This formulation allows the use of better numerical solvers for larger step sizes and fewer model iterations. Under the SDE framework, previous works alter the numerical solver \cite{song2021score}, the initialization point \cite{chung2022come}, the step-size \cite{jolicoeur2021gotta}, and the order of the solver \cite{karras2022elucidating}. The SDE can also be reformulated as an ordinary differential equation which is deterministic and therefore easier to accelerate. Many works \cite{liu2021pseudo, lu2022dpm, lu2022dpms, zhang2022fast, karras2022elucidating} hence built upon the ODE formulation and propose better ODE solvers for the problem. Although existing studies have extensively explored how a better numerical solver can be used to accelerate diffusion inference. They have not taken the sensitivity of ODEs into consideration.

\section{Background}

\subsection{Diffusion Models}

Assuming data follow an unknown true distribution $p(\x)$, diffusion models \cite{sohl2015deep, ho2020denoising, song2021score, kingma2021variational} define a generation process. For any $p(\x)$, diffusion models learn a denoising process by iteratively adding noise to original data (denoted as $\x_0$) from time step $0$ until time step $T$. The forward process adds noise to $\x_0$ such that at time step $t$, the distribution of $\x_t$ conditioned on $\x_0$ is
\begin{equation}\small
    q(\x_t|\x_0)=\mathcal{N}\left (\alpha(t)x_0,\sigma^2(t)\bm{I}\right ),
\end{equation}

where $\alpha(t),\sigma(t)$ are real-valued positive functions with bounded derivatives. The signal-to-noise ratio (SNR) is defined as $\gamma=\alpha^2(t)/\sigma^2(t)$. By making the SNR decreasing, the marginal distribution of $\x_T$ approximates a zero-mean Gaussian, i.e.
\begin{equation}\small
    q(\x_T) \approx \mathcal{N}\left (\bm{0}, \tilde{\sigma}^2\bm{I} \right ).
\end{equation}

The noise-adding forward process transforms a data sample from the original distribution to the zero-mean Gaussian $q(\x_T)$, while the backward process randomly samples from $q(\x_T)$ and denoises the sample with a neural network parameterized conditional distribution $q(\x_s | \x_t)$ where $s<t$ until it reaches time step $0$.

\subsection{The Differential Equation Formulation}

\citet{kingma2021variational} proves that the forward process is equivalent (in distribution) to the following stochastic differential equation (SDE) in terms of the conditional distribution $q(\x_t|\x_0)$,
\begin{align}\small
    \d \x_t&=f(t)\x_t \d t + g(t)\d \bm{w},\\
    f(t)&=\frac{\d \log \alpha(t)}{\d t},\\
    g(t)&=\frac{\d \sigma^2(t)}{\d t}+2\frac{\d\log \alpha(t)}{\d t}\sigma^2(t),
\end{align}
where $\bm{w}$ is the standard Wiener process.

\citet{song2021score} shows that the forward SDE has an equivalent reverse SDE starting from time $T$ with the marginal distribution $q(\x_T)$,
\begin{equation}\small
    \d \x_t=[f(t)\x_t-g^2(t)\nabla_\x \log q_t(\x_t)]\d t+g(t)\d \bar{\bm{w}}_t, \label{eq:revsde}
\end{equation}
where $\bar{\bm{w}}_t$ is the reverse Wiener process and $t$ goes from $0$ to $T$.

In practice, $\nabla_\x \log q_t(\x_t)\approx \eps / \sigma(t)$ is estimated using a noise-predictor function $\bm{\epsilon}_\theta(\x_t,t)$, where $\eps$ is the Gaussian noise added to $\x_0$ to obtain $\x_t$, i.e.,

\begin{equation}\small
\x_t=\alpha(t)\x_0+\sigma(t)\eps,\quad\epsilon\sim\mathcal{N}(\bm{0},\bm{I}).
\end{equation}

To learn $\eps_\theta$, the following objective function is minimized \cite{ho2020denoising, song2021score},

\begin{equation}\small
L(\theta)=\int_0^T\omega(t)\E_{q(\x_0)}\E_{q(\x_t|\x_0)}\left[\left\Vert\eps_\theta(\x_t,t)-\eps\right\Vert^2\right]\d t,
\end{equation}

where $\omega(t)$ is a weighting function, and the integral is estimated using random samples.

\citet{song2021score} proves that the following ordinary differential equation (ODE) is equivalent to \autoref{eq:revsde},
\begin{equation}\small
    \frac{\d \x_t}{\d t}=\left [f(t)\x_t-\frac{1}{2}g^2(t)\nabla_\x \log q_t(\x_t)\right ]. \label{eq:revode}    
\end{equation}
When $\nabla_\x\log q_t(\x_t)$ is replaced by its estimation, we obtain the neural network parameterized ODE,
\begin{equation}\small
    \frac{\d \x_t}{\d t}=\left [f(t)\x_t-\frac{g^2(t)}{2\sigma^2(t)}\eps_\theta (\x_t,t)\right ]. \label{eq:node}    
\end{equation}

The inference process of diffusion models can be formulated as solving \autoref{eq:node} given a random sample from the Gaussian distribution $q(\x_T)$. For each iteration in solving the equation, the noise-predictor function $\eps_\theta(\x_t,t)$ is estimated with the trained neural network $\theta$. Therefore, the inference time consumption can be approximated by the number of function estimations (NFEs), i.e., how many times $\eps_\theta(\x_t,t)$ is estimated.

\section{The \method~Method}

\newcommand{\algorithmautorefname}{Algorithm}

Given a starting sample $\x_T$, a guidance signal $y$, ReDi accelerates diffusion inference by skipping some intermediate samples in the sample trajectory to reduce NFEs.

\begin{figure*}[th!]
\vskip 0.1in
\begin{center}
\centerline{\includegraphics[width=\textwidth]{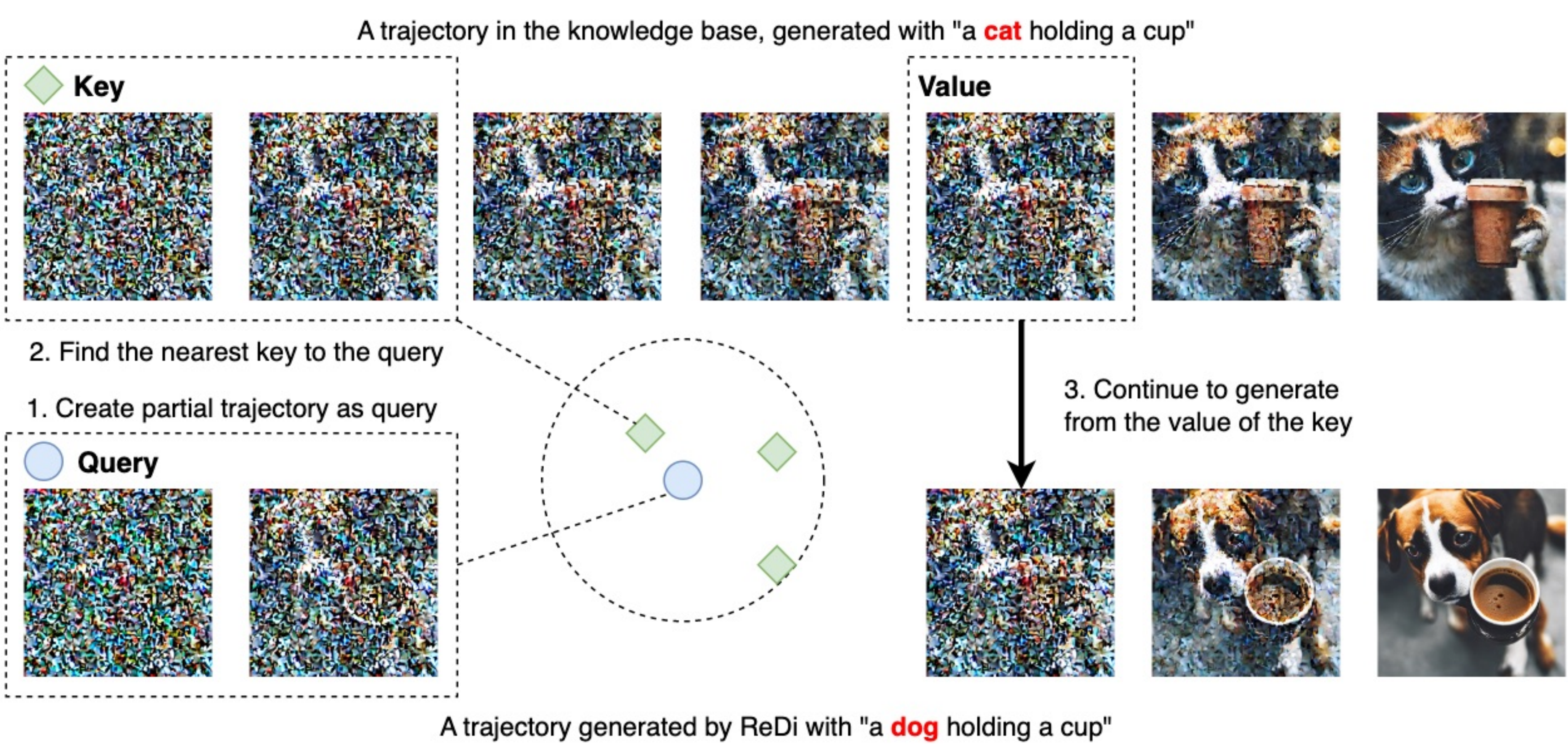}}
\caption{The inference procedure of \textsc{ReDi}. The upper part is a complete trajectory generated by Stable Diffusion to build the knowledge base. The lower part is a trajectory generated by \textsc{ReDi} with some intermediate steps skipped.}
\label{main_arch}
\end{center}
\vskip -0.1in
\end{figure*}

\subsection{The Sample Trajectory}

\begin{definition}
\label{def:traj}
Given a starting sample $\x_T$, the \textit{sample trajectory} of a diffusion model is a sequence of intermediate samples generated in the iterative process from $\x_T$ to $\x_0$. For a particular time step size $\delta$, the sample trajectory is the sequence $\{\x_T,\x_{T-\delta}, \x_{T-2\delta},\dots,\x_\delta, \x_0\}$, where $T$ is the initial time step.
\end{definition}

The sample trajectory describes the intermediate steps to generate the final data sample (e.g. an image), so the inference time linearly correlates with its length, i.e., the number of estimations used. While previous numerical solvers work towards enlarging the step size $\delta$, \textsc{ReDi} aims at skipping some intermediate steps to reduce the length of the trajectory. \textsc{ReDi} is able to do so because the first few steps determine only the layout of the image which can be shared by many, and the following steps determine the details \cite{Meng2021SDEditGI, wu2022uncovering}.

In the following section, we describe the \textsc{ReDi} algorithm with this definition.

\begin{algorithm}[h]
\caption{\textsc{ReDi} Knowledge Base Construction}\label{alg:conskb}
 \small
\begin{algorithmic}
\INPUT 
A dataset $\mathcal{D}=\{(\x^{(i)},y^{(i)})\}_{i=1}^N$ where $y^{(i)}$ is some guidance signal and $\x^{(i)}$ is a data sample;  \\
A pretrained diffusion model $\theta$ that computes $\eps_\theta(\x_t,t, y)$, the noise estimation of $\x_t$ at time step $t$; \\
Two constant time steps $k$ and $v$, where $k > v$; \\
A numerical sampler $S$, $\x_{t-1}=S(\x_t,t, \theta)$.

\OUTPUT 
A knowledge base $\mathcal{B}=\{(key^{(i)},val^{(i)})\}_{i=1}^N$
\\

\FOR {$i \gets 1 \textbf{ to } N$}
\STATE $\x_T\sim p(x_T|\x^{(i)})$

\upd{// Note that the signal-to-noise ratio is close to 0 at time step $T$ due to the noise ratio of diffusion models. Therefore sampling $\x_T$ from $p(\x_T|\x^{(i)})$ is almost the same as from $p(\x_T)$.}

\upd{// More discussions about the initial sample in Appendix \ref{sec:initial_sample}.}
    \FOR {$j \gets T \textbf{ downto } 1$}
        \STATE $ \x_{j-1}\gets S_\theta\left(\x_j,j,y^{(i)}\right)$
    \ENDFOR\  // Calculating the trajectories of the $i$th sample
    \STATE $key^{(i)}\gets \x_{k}$ // An early of the trajectory as the key
    \STATE $val^{(i)}\gets \x_{v}$ // An intermediate sample as the value
\ENDFOR

\STATE \textbf{return }$\{(key^{(i)},val^{(i)})\}_{i=1}^N$

\end{algorithmic}
\end{algorithm}

\subsection{The \textsc{ReDi} Inference framework} \label{sec:framework}

Given a trained diffusion model, \textsc{ReDi} does not require any change to the parameters. It only needs access to the sample trajectory of generated images. We show the inference procedure of \textsc{ReDi} in \autoref{main_arch}. Instead of generating all $T/\delta$ intermediate samples in the trajectory, \textsc{ReDi} skips some of them to reduce number of function estimations (NFEs) and improve efficiency. This is done by generating the first few samples $\x_{T..k}$, using them as a query to retrieve a similar trajectory $\x'_{T..k}$, and then starting from $\x'_v$ of the trajectory retrieved. This way, the NFEs spent to go from time step $k$ to time step $v$ would be unnecessary.

As shown in \autoref{main_arch}, the retrieval of the similar trajectory $\x'$ depends on a precomputed knowledge base $\mathcal{B}$. We formally describe the construction of $\mathcal{B}$ in \autoref{alg:conskb}. \textsc{ReDi} first computes the sample trajectories for the data samples in a dataset $\mathcal{D}$. For every sample trajectory $\{\x_{n\delta}\}_{n=T/\delta,\dots,1,0}$, a sample early in generation $\x_{k}$ is chosen as the key, while a later sample $\x_v$ is chosen as the value. The time consumption for computing one key-value pair is similar to that of one generation of the model. The total time consumption is proportional to $|\mathcal{D}|$. With a moderate-sized $\mathcal{D}$, not only can we achieve comparable performance with much less time, we can also perform zero-shot domain adaptation.

The inference process of \textsc{ReDi} is formally described in \autoref{alg:sample}. Given a guidance signal $y$ (in our case, a text prompt), we want to generate a data sample (in our case, an image) $\x$ from it. We first generate the first few steps $\x_{T..k}$ to query the knowledge base $\mathcal{B}$ with $\x_k$. We find the top-$H$ keys that are closest to $q$ in terms of the distance measure $d$. Then we find out the set of weights $w^*$ that make the linear combination of the top-$H$ keys approximate $q$ the best. With $w^*$, we linearly combine the value and use that combination as the initialization point for the remaining steps of the sampling process.

\begin{algorithm}[t]
\caption{\textsc{ReDi} Inference}\label{alg:sample}
\small
\begin{algorithmic}
\INPUT
$\theta,k,v,S$ are the same as defined in \autoref{alg:conskb};\\
$\mathcal{B}$ is the knowledge base computed by \autoref{alg:conskb};\\
The guidance signal $y$;\\
The number of nearest neighbors $H$ we want to retrieve;\\
A distance measure $d(\cdot, \cdot)$ between a query and a key.

\OUTPUT A data sample $x$ conditioned on the guidance signal $y$.
\\

\STATE $\x_T\sim p(\x_T)$
\FOR {$i\gets T \textbf{ to } k+1$}
    \STATE $\x_{i-1}\gets S_\theta(\x_i,i,y)$
\ENDFOR
\STATE $q\gets \x_{k}$ // Computing the first few samples as the query
\STATE Find the top-$H$ neighbors $j_1,j_2,\dots,j_H$ from $\mathcal{B}$ that are nearest to $q$ (measured by $d$)
\STATE $w^*\gets \arg\min_{w}d\left(q,\sum_{i=1}^Hw_ikey^{(j_i)}\right)$
\STATE $\hat x_{v}\gets \sum_{i=1}^Hw^{*}_ival^{(j_i)}$ \\
\FOR {$i\gets v \textbf{ downto } 1$}
    \STATE $\hat x_{i-1}\gets S_\theta(\hat x_i,i,y)$
\ENDFOR
\STATE \textbf{return } $\hat x_0$
\end{algorithmic}
\end{algorithm}

\subsection{Extending \textsc{ReDi} for Zero-Shot Domain Adaptation} \label{sec:doma}

One limitation of the \textsc{ReDi} framework described in \autoref{sec:framework} is its generalizability. When the guidance signal $y$ contains out-of-domain information that does not exist in $\mathcal{B}$, it is difficult to retrieve a similar trajectory from $\mathcal{B}$ and run \textsc{ReDi}. Therefore, We propose an extension to \textsc{ReDi} in order to generalize it to out-of-domain guidance. For an out-of-domain guidance signal $y$, we break it into 2 parts - the domain-agnostic $y^{\text{in}}$, and the domain-specific $y^{\text{out}}$. We use $y^{\text{in}}$ to generate the partial trajectory as the retrieval key. After retrieval, we start from the retrieved value with both $y^{\text{in}}$ and $y^{\text{out}}$ as guidance signal.

For example, under the image synthesis setting, when $\mathcal{B}$ contains style-free images that are generated without any style specifier in the prompt, it is difficult for \textsc{ReDi} to synthesize images from a stylistic prompt $y$ because finding a stylistic trajectory from a style-free knowledge base is hard. However, with the proposed extension, \textsc{ReDi} is able to synthesize stylistic images.

When a stylistic guidance signal $y$ is given, the part of $y$ describing the content of the image is the domain-agnostic $y^{\text{in}}$, and the part of $y$ specifying the style of the image is the domain-specific $y^{\text{out}}$. Although it is difficult to find a trajectory similar to one generated by $y=(y^{\text{in}},y^{\text{out}})$, it is feasible to retrieve a trajectory similar to one generated by $y^{\text{in}}$. Therefore, we first use the content description to generate the retrieval key and then use the whole prompt for the following sampling steps to make the image stylistic.

\section{Theoretical Analysis}

\newcommand{\algorithmautorefname}{Algorithm}
\label{sec:proof}

We analyze in this section whether \textsc{ReDi} is guaranteed to work. Our theorem is based on two assumptions.

\begin{assumption}
The noise predictor model $\eps_\theta(\x_t,t)$ is $L_0$-Lipschitz.
\end{assumption}

This assumption is used in \citet{lu2022dpm} to prove the convergence of DPM-solver. \citet{salmona2022can} also argues that diffusion models are more capable of fitting multimodal distributions than other generative models like VAEs and GANs because of its small Lipschitz constant. The fact that $\eps$ is an estimate of the Gaussian noise added to the original image suggests that a small perturbation in $\x_t$ does not change the noise estimation very much.

\begin{assumption}
The distance between the query and the nearest retrieved key is bounded, i.e. $d(q, \text{key}) \le \varepsilon$.
\end{assumption}

This assumes that the nearest neighbor that \textsc{ReDi} retrieves is ``near enough'', which we show empirically in \autoref{sec:exp}. 

Given these assumptions, we can prove a theorem saying the distance between the retrieved value and the true sample generated retrieved value is an estimate near enough to the actual $\x_v$.

\begin{theorem} \label{thm:main}
If $d(\x_k, \text{key})\le \varepsilon$, then $d(\x_v, \text{val}) \le e^{\mathcal{O}(k-v)}\varepsilon$.
\end{theorem}

Here, $x_k$ is the generated early sample used to retrieve from the knowledge base. $\text{key}$ is the $k$-th sample from a trajectory stored in the knowledge base. $\text{val}$ is the $v$-th sample from the same trajectory as $\text{key}$. $\x_v$ is the true $v$-th sample if we generate the full trajectory using the original diffusion sampler starting from $\x_k$.

\textit{Proof: } We first define $\bm{h} (\x,t):=\dfrac{\d \x_t}{\d t}$ and prove it's Lipschitz continuous. This is equivalent to proving $\left |\dfrac{\partial \bm{h}}{\partial \x}\right |$ is bounded:

\begin{align}
\left |\dfrac{\partial \bm{h}}{\partial \x}\right |&=\left|f(t)-\frac{g^2(t)}{2\sigma(t)}\frac{\partial \eps}{\partial \x}\right| \le
\left |f(t)\right |+\left|\frac{g^2(t)}{2\sigma(t)}\right|\left |\dfrac{\partial \eps}{\partial \x}\right |\\
&\le \left |f(t)\right |+\left|\frac{g^2(t)}{2\sigma(t)}\right|L_0. \label{eq:lip}
\end{align}

Note that \autoref{eq:lip} follows from the Lipschitz continuity of $\eps$. $\left |f(t)\right |+\left|\frac{g^2(t)}{2\sigma(t)}\right|L_0$ is bounded by the bounds of $f$ and $g$, which is determined by the range of $t$.

Since $\dfrac{\d \x_t}{\d t}$ is $L$-Lipschitz, the sensitivity of ODE \cite{khalil2002nonlinear} suggests that for any two solutions $\x$ and $\y$ to \autoref{eq:node} satisfies
\begin{equation}\label{eq:ode}
    |\x_v-\y_v|\le e^{L(k-v)}|\x_k-\y_k|.
\end{equation}

We summarize the proof of ODE sensitivity in Appendix \ref{sec:a_proof} to provide a more thorough perspective to our proof. With \ref{eq:ode}, the theorem is proven because $key$ and $val$ are $y_v$ and $y_k$ from a solution to \autoref{eq:node} according to \autoref{alg:conskb}. $\square$

This theorem can be interpreted as that if the retrieved trajectory is near enough, the retrieved $\x'_v$ would be a good-enough surrogate for the actual $\x_V$. Note that multiple factors affect the proven bound, namely the Lipschitz constant $L$, the distance to the retrieved trajectory $d$, and the choice of key and value steps $k$ and $v$.

\section{Experiments} \label{sec:exp}

This section presents experiments for \textsc{ReDi} applied in both the inference acceleration setting and the stylization setting. We conduct these experiments to investigate the following research questions:
\begin{itemize}[noitemsep, leftmargin=*, topsep=2pt]
    \item Is \textsc{ReDi} capable of keeping the generation quality while speeding up the inference process?
    \item Is the sample trajectory a better key for retrieval-based generation than text-image representations?
    \item Is \textsc{ReDi} able to generate data from a new domain without re-constructing the knowledge base?
\end{itemize}

We conduct our experiments on Stable Diffusion v1.5\footnote{https://github.com/CompVis/stable-diffusion}, an implementation of latent diffusion models \cite{rombach2022high} with 1000M parameters. Our inference code is based on Huggingface Diffusers\footnote{https://github.com/huggingface/diffusers}. To show that \textsc{ReDi} is orthogonal to the choice of numerical solver, we conduct experiments with two widely used numerical solvers - pseudo numerical methods (PNDM) \cite{liu2021pseudo} and multistep DPM solvers \cite{lu2022dpm, lu2022dpms}. To obtain the top-$H$ neighbors, we use ScaNN \cite{avq_2020}, which adds a neglectable overhead to the generation process for retrieving the nearest neighbors.

\subsection{\textsc{ReDi} is efficient in precompute and inference} \label{sec:exp1}
\begin{table}[t]
\caption{The FID scores $\downarrow$ of \textsc{ReDi} when it's applied to PNDM. With 20 NFEs, \textsc{ReDi} is able to achieve better quality than the 40-step PNDM, making the inference process 2$\times$ faster. The NFE of the 20-step \textsc{ReDi} is comparable to a 1000-step DDIM sampler.}
\label{tab:fid_pndm}
\vskip 0.15in
\begin{center}
\begin{sc}
\begin{tabular}{lllll}
    \hline
           & \multicolumn{4}{c}{\begin{tabular}[c]{@{}@{}c@{}}NFEs\end{tabular}} \\
Sampler    & 20                              & 30                              & 40        & 1000                      \\ \hline DDIM & - & - & - & 0.255\\ 
PNDM       & 0.274                           & 0.268                           & 0.262           & -                \\
\quad +ReDi, H=1 & 0.265                           & 0.264                           & \textbf{0.262}       & -                    \\
\quad +ReDi, H=2 & \textbf{0.255}                           & \textbf{0.247}                           & \textbf{0.252}           & -                \\ \hline
\end{tabular}
\end{sc}
\end{center}
\end{table}

\begin{figure}[h]
\vskip 0.1in
\begin{center}
\centerline{\includegraphics[width=0.45\textwidth]{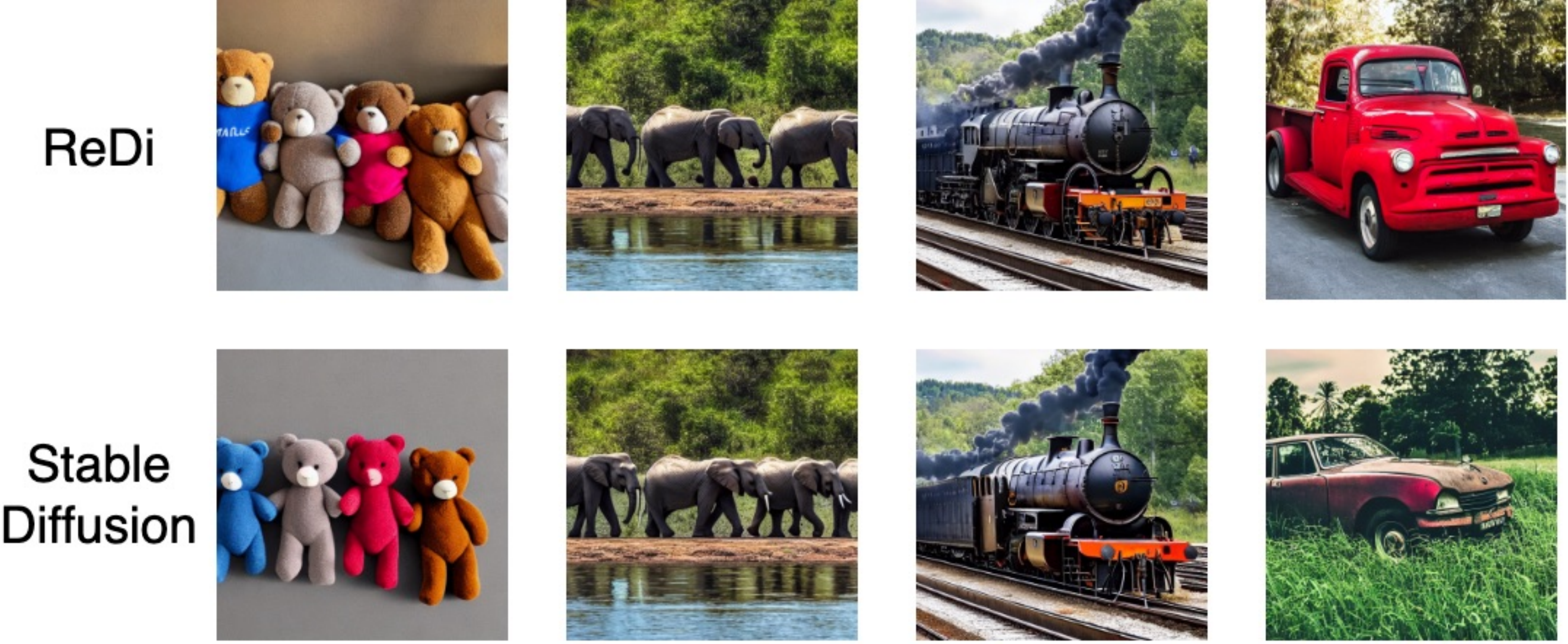}}
\caption{The image samples generated by Stable Diffusion and \textsc{ReDi} with the same set of prompts.}
\label{fig:redisamples}
\end{center}
\vskip -0.1in
\end{figure}

To show \textsc{ReDi}'s ability to sample from trained diffusion models efficiently, we use \textsc{ReDi} to sample from Stable Diffusion, with two different numerical solvers, PNDM and DPM-Solver. We build the knowledge base $\mathcal{B}$ upon MS-COCO \cite{lin2014microsoft} training set (with 82k image-caption pairs) and evaluate the generation quality on 4k random samples from the MS-COCO validation set. We use the Fréchet inception distance (FID) \cite{heusel2017gans} of the generated images and the real images. In practice, we choose $L^2$-norm as our distance measure $d$ and calculate the optimal combination of neighbors using least square estimation.

Note that although the training of Stable Diffusion costs as much as 17 GPU years on NVIDIA A100 GPUs, constructing $\mathcal{B}$ with PNDM only requires 1 GPU day, which is much more efficient compared to learning-based methods like progressive distillation \cite{meng2022distillation}.

For PNDM, we generate 50-sample trajectories to build the knowledge base. We choose the key step $k$ to be $40$, making the first 10 samples in the trajectory the key, and alternate the value step $v$ from $30$ to $10$. Different combinations of key and value steps determine how many NFEs are reduced. For DPM-solver, we choose the length of the trajectory to be $20$ and conduct experiments on $k=5, v=8/10/13/15$. To compare the performance of \textsc{ReDi} with the existing numerical solvers. We use them to generate images with the same NFE budget and compare their FIDs with \textsc{ReDi}.

We show the results of our experiments in \autoref{tab:fid_pndm} and \autoref{tab:fid_dpm}. Some samples generated by ReDi are listed in \autoref{fig:redisamples}. For both PNDM and DPM-Solver, we report the FID scores of the sampler without \textsc{ReDi}, \textsc{ReDi} with top-1 retrieval, and \textsc{ReDi} with top-2 retrieval. The results indicate that with the same number of NFEs, \textsc{ReDi} is able to generate images with a better FID. They also indicate that when \textsc{ReDi} is combined with numerical solvers, we can achieve better performance with only 40\% to 50\% of the NFEs. \upd{In terms of clock time, 50-step PNDM takes 2.94 seconds, while the 30-step \textsc{ReDi} takes 1.75 seconds to generate one sample. The precomputation time for building the vector retrieval index and retrieving top-$h$ neighbors, when amortized by 4000 inferences, is 0.0077 seconds, which only takes 0.4\% of the inference time.} Therefore, \textsc{ReDi} is capable of keeping the generation quality while improving inference efficiency.

\upd{We further validate the effectiveness of \method~with more ablation studies and baselines in Appendix \ref{sec:effectiveness}.}

\begin{table}[t]
\caption{The FID scores of \textsc{ReDi} when it's applied to DPM-Solver.}
\label{tab:fid_dpm}
\vskip 0.15in
\begin{center}
\begin{sc}
\begin{tabular}{lllll}
    \hline
           & \multicolumn{4}{c}{\begin{tabular}[c]{@{}@{}c@{}}NFEs\end{tabular}} \\
Sampler    & 10                      & 12                      & 15                     & 17                     \\ \hline
DPM-Solver & 0.268                   & 0.265                   & 0.263                  & 0.258                  \\
\quad+ReDi, H=1 & 0.265                   & 0.259                   & 0.256                  & 0.255                  \\
\quad+ReDi, H=2 & \textbf{0.261}                   & \textbf{0.257}                   & \textbf{0.253}                  & \textbf{0.252}                  \\ \hline
\end{tabular}
\end{sc}
\end{center}
\end{table}

\subsection{\upd{Early samples from} trajectories are better retrieval keys than text-image representations}

One major difference between \textsc{ReDi} and previous retrieval-based samplers is the keys we use. Instead of using text-image representations like CLIP embeddings, we use an early step from the sample trajectory as the key. Theoretically, this is the optimal solution, because we are directly minimalizing the bound from \ref{thm:main}\ref{thm:main} by minimalizing $|\x_k-\y_k|$. In this section, we show empirically that sample trajectory is indeed the better retrieval key.

We compare CLIP embeddings and trajectories by using them alternatively as the key. To use CLIP embeddings as key, we build a knowledge base similar to the one we use in \autoref{sec:exp1} with them, where the keys are CLIP embeddings and the values are later samples in the sample trajectory. We run the same inference process except with a different knowledge base.

To evaluate the performance of retrieval keys, we use two criteria - the distance between the actual value and the retrieved value $d(\x_v, \hat\x_v)$, and the FID scores of the generated images. We report the results in \autoref{tab:keycomp}. Under all NFE settings, the trajectories serve a better purpose as the key for \textsc{ReDi} than CLIP embeddings. This corresponds well to \autoref{thm:main}.

\begin{figure*}[t!]
     \centering
     \begin{subfigure}[b]{0.3\textwidth}
         \centering
         \includegraphics[width=\textwidth]{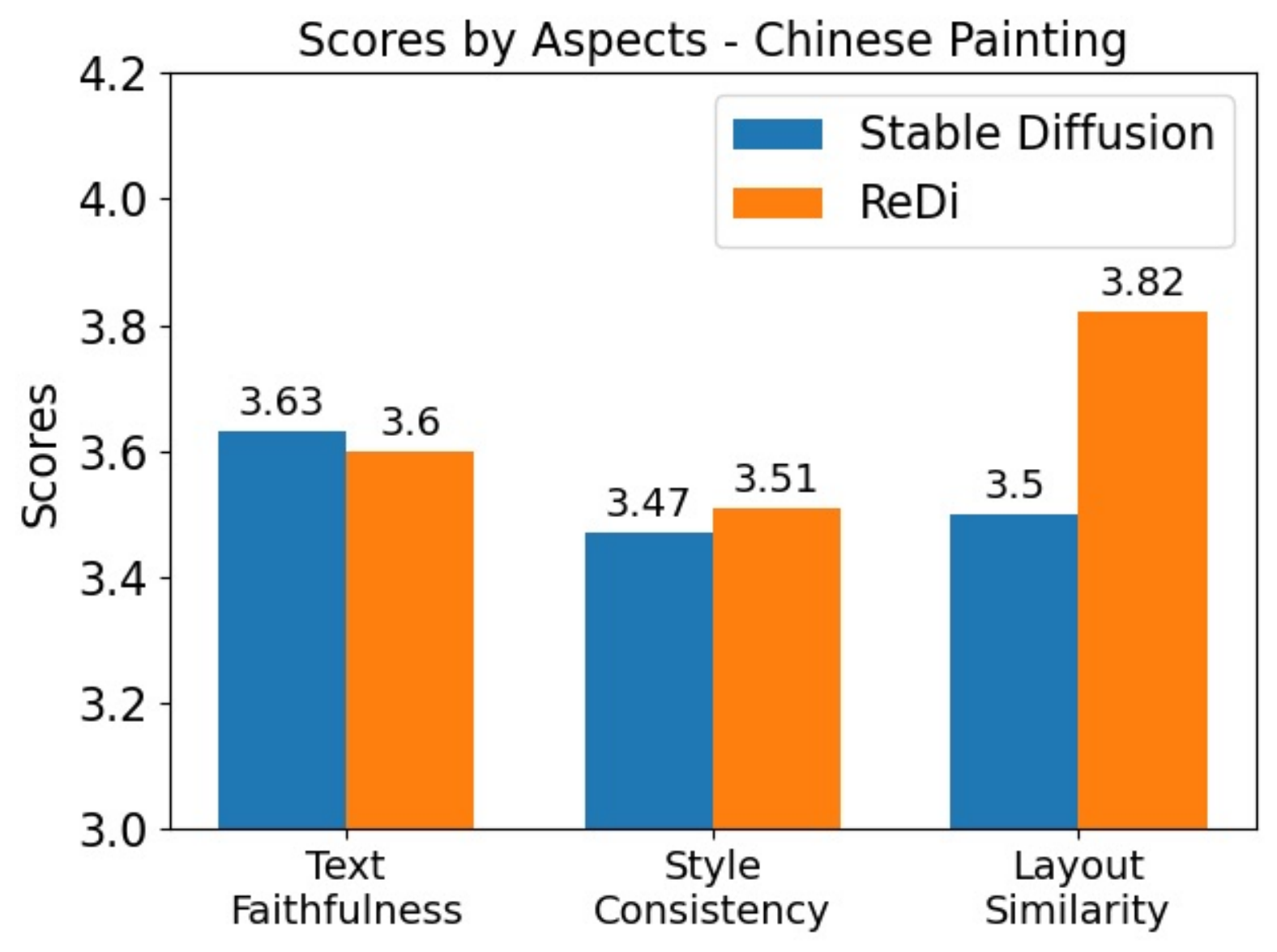}
         \label{fig:chinese_style}
     \end{subfigure}
     \hfill
     \begin{subfigure}[b]{0.3\textwidth}
         \centering
         \includegraphics[width=\textwidth]{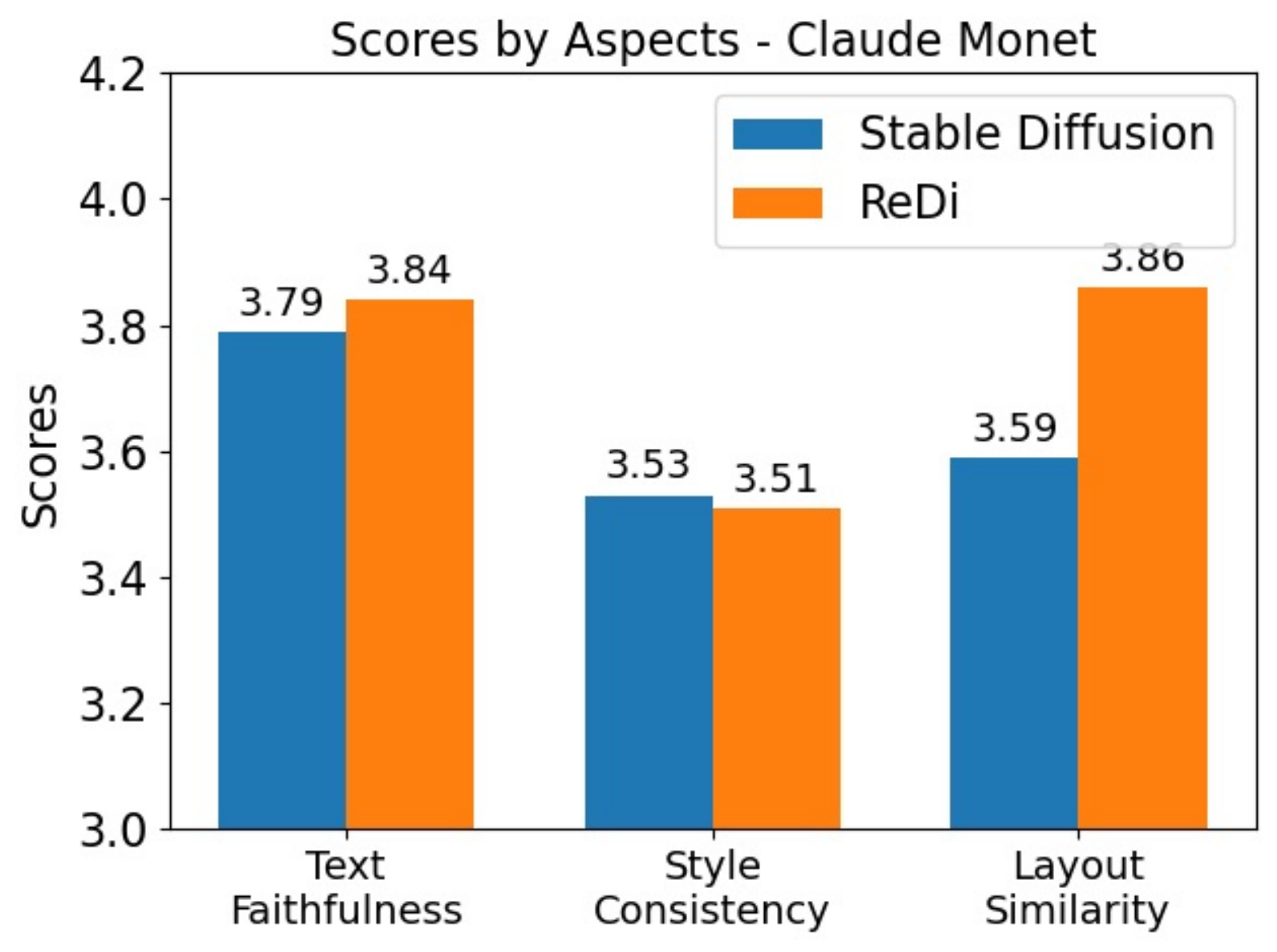}
         \label{fig:monet_style}
     \end{subfigure}
     \hfill
     \begin{subfigure}[b]{0.3\textwidth}
         \centering
         \includegraphics[width=\textwidth]{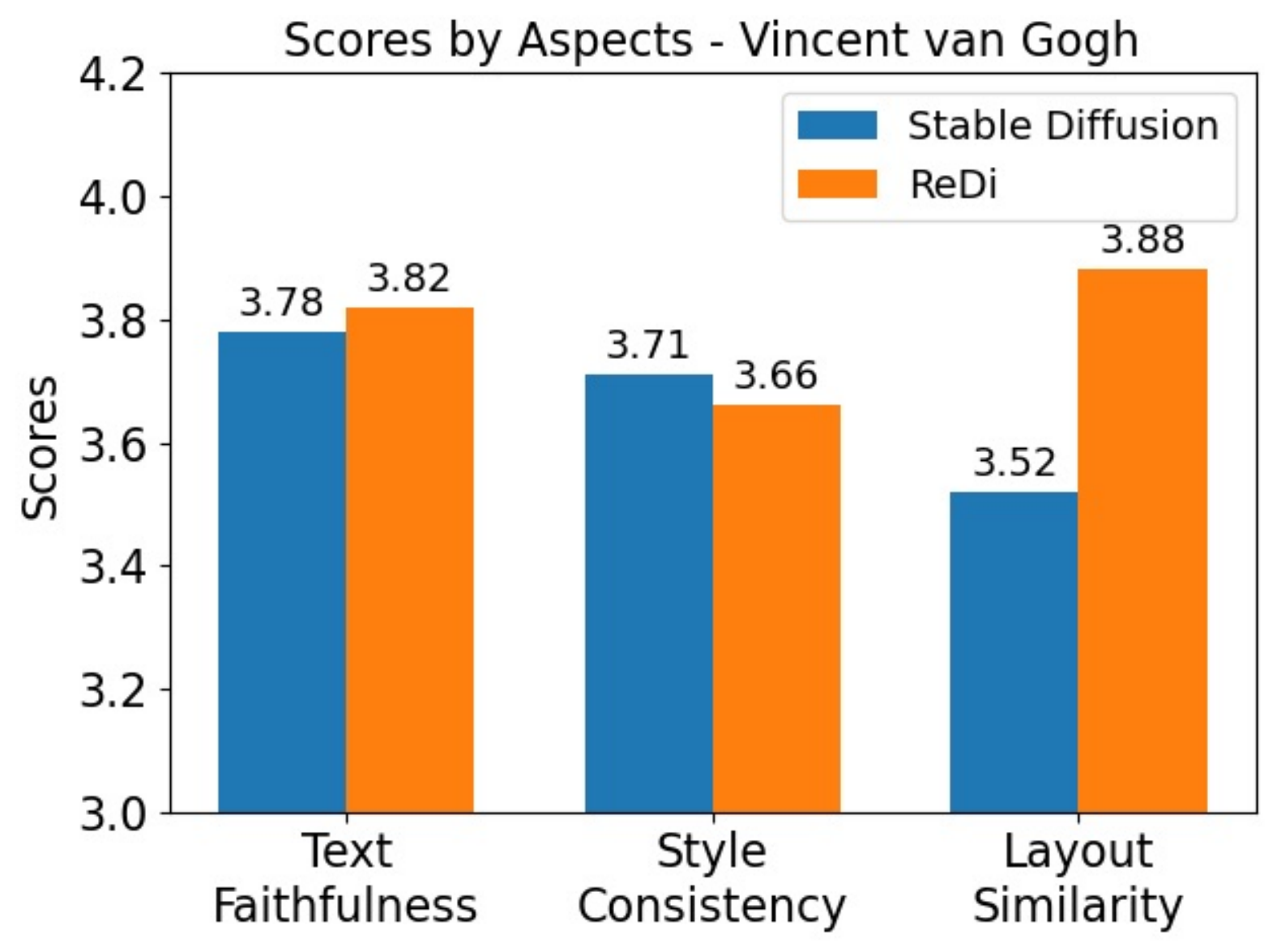}
         \label{fig:vincent_style}
     \end{subfigure}
        \caption{Textual faithfulness, style consistency, and layout similarity scores for Stable Diffusion and ReDi, in 3 different styles. In terms of layout similarity, ReDi is significantly better than Stable Diffusion, indicating that ReDi is able to control the style without changing the layout.}
        \label{fig:fid_style}
\end{figure*}

\begin{table}[t]
\caption{The L2 distance and FID scores of using FID and trajectory as the retrieval key.}
\label{tab:keycomp}
\vskip 0.15in
\begin{center}
\begin{sc}
\begin{tabular}{llll}
\hline
NFE                 & Key        & L2 norm ↓ & FID ↓  \\ \hline
\multirow{2}{*}{40} & CLIP       & 9.03      & 0.2709 \\
                    & Trajectory & 7.95      & 0.2626 \\ \hline
\multirow{2}{*}{30} & CLIP       & 8.53      & 0.2784 \\
                    & Trajectory & 7.28      & 0.2643 \\ \hline
\end{tabular}
\end{sc}
\end{center}
\end{table}

\subsection{\textsc{ReDi} can perform zero-shot domain adaptation without a domain-specific knowledge base}

We use the extended \textsc{ReDi} framework from \autoref{sec:doma} to generate domain-specific images. In particular, we conduct experiments on image stylization~\cite{Fu2022LanguageDrivenAS,Feng2022TrainingFreeSD} with the style-free knowledge base $\mathcal{B}$ from \ref{sec:exp1}. To generate an image with a specific style, we do not build a knowledge base from stylistic images. Instead, we use the content description $y^{\text{content}}$ to generate the partial trajectory as key. After retrieval, we change the prompt from $y^{\text{content}}$ to the combination of $y^{\text{content}}$ and $y^{\text{style}}$, where $y^{\text{style}}$ is the style description in the prompt.

In our experiments, we transfer the validation set of MS COCO to three different styles. The content description $y^{\text{content}}$ directly comes from the image captions, while the style description $y^{\text{style}}$ is appended as a suffix to the prompt. The style descriptions for the three styles are ``\textit{a Chinese painting}'', ``\textit{by Claude Monet}'', and ``\textit{by Vincent van Gogh}''.

Unlike other experiments, in this experiment, we choose $K=47$ and $V=40$. Because from the preliminary experiments, we find that the style of the image is determined much earlier than its detailed content. For 100 randomly sampled captions from MS COCO validation, we generate the corresponding images in all three styles. To evaluate \textsc{ReDi}'s performance, we compare them with images generated by the original Stable Diffusion with PNDM solver.

We asked human evaluators from Amazon MTurk to evaluate the generated images. They are paid more than the local minimum wage. Every generated image is rated in three aspects, text faithfulness, style consistency, and layout similarity. Every aspect is rated on a scale of 1 to 5 where 5 stands for the highest level. Textual faithfulness represents how faithful the image is depicting the content description. Style consistency represents how consistent the image is to the specified style. Layout similarity represents how similar the layout of the stylistic image is to the layout of the style-free image.

\begin{figure}[H]
\vskip 0.1in
\begin{center}
\centerline{\includegraphics[width=0.5\textwidth]{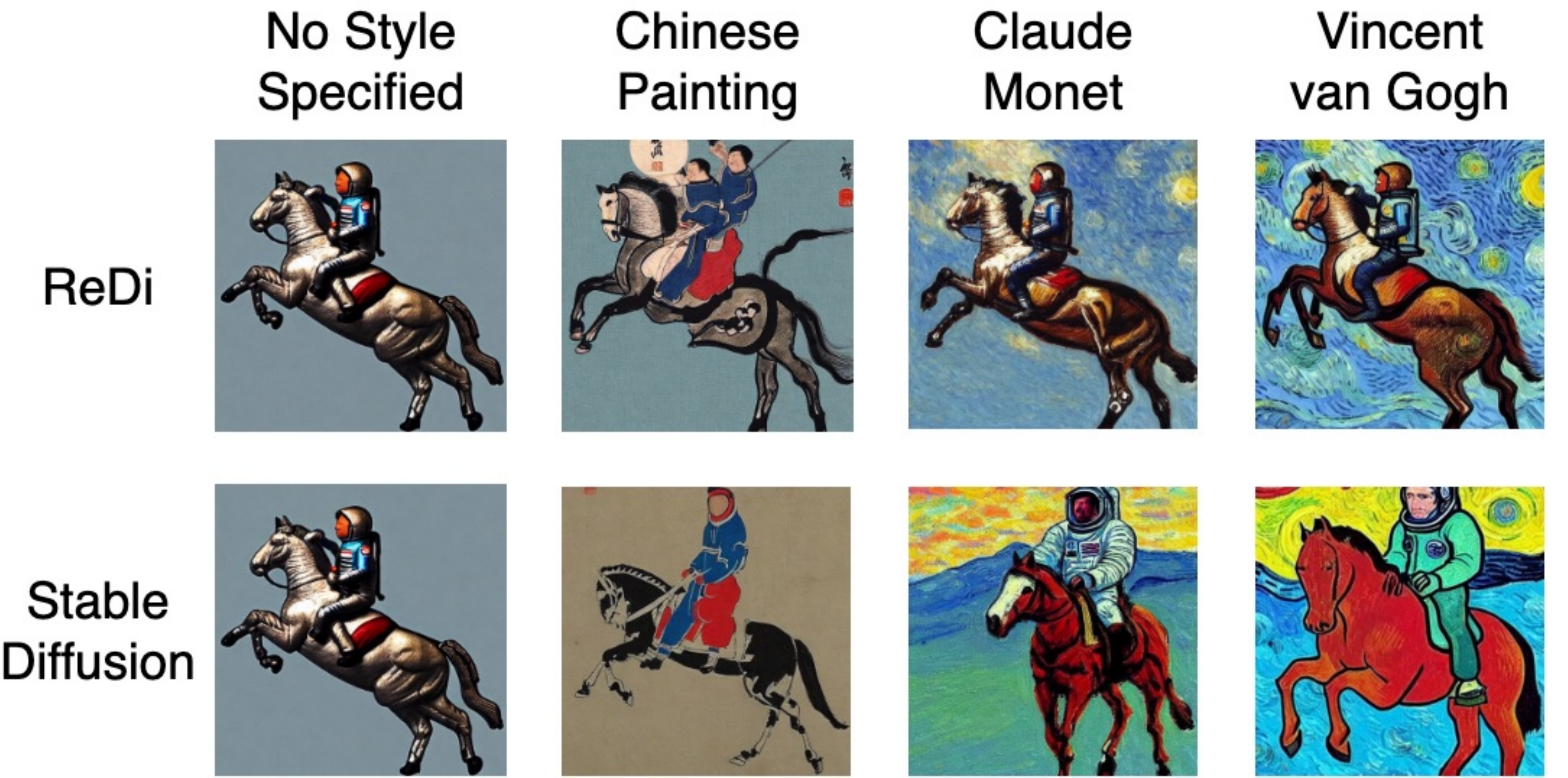}}
\caption{The image samples generated by Stable Diffusion and \textsc{ReDi} with the prompt ``\textit{an astronaut riding a horse}'' in different styles. Without extra precomputation, \textsc{ReDi} is able to control the style and keep the same layout, while speeding up the inference.}
\label{fig:samples}
\end{center}
\vskip -0.1in
\end{figure}

We report the results of the human evaluation of the 3 different styles in \autoref{fig:fid_style}. In terms of text faithfulness and style, \textsc{ReDi}'s evaluation is comparable to Stable Diffusion. In terms of layout similarity, the images generated by \textsc{ReDi} have significantly more similar layouts. This indicates that by retrieving the early sample in the trajectory, \textsc{ReDi} is able to keep the layout unchanged while transferring the style. This finding is also demonstrated by the qualitative examples in \autoref{fig:samples}.%

\begin{figure*}[ht]
     \centering
     \begin{subfigure}[b]{0.3\textwidth}
         \centering
         \includegraphics[width=\textwidth]{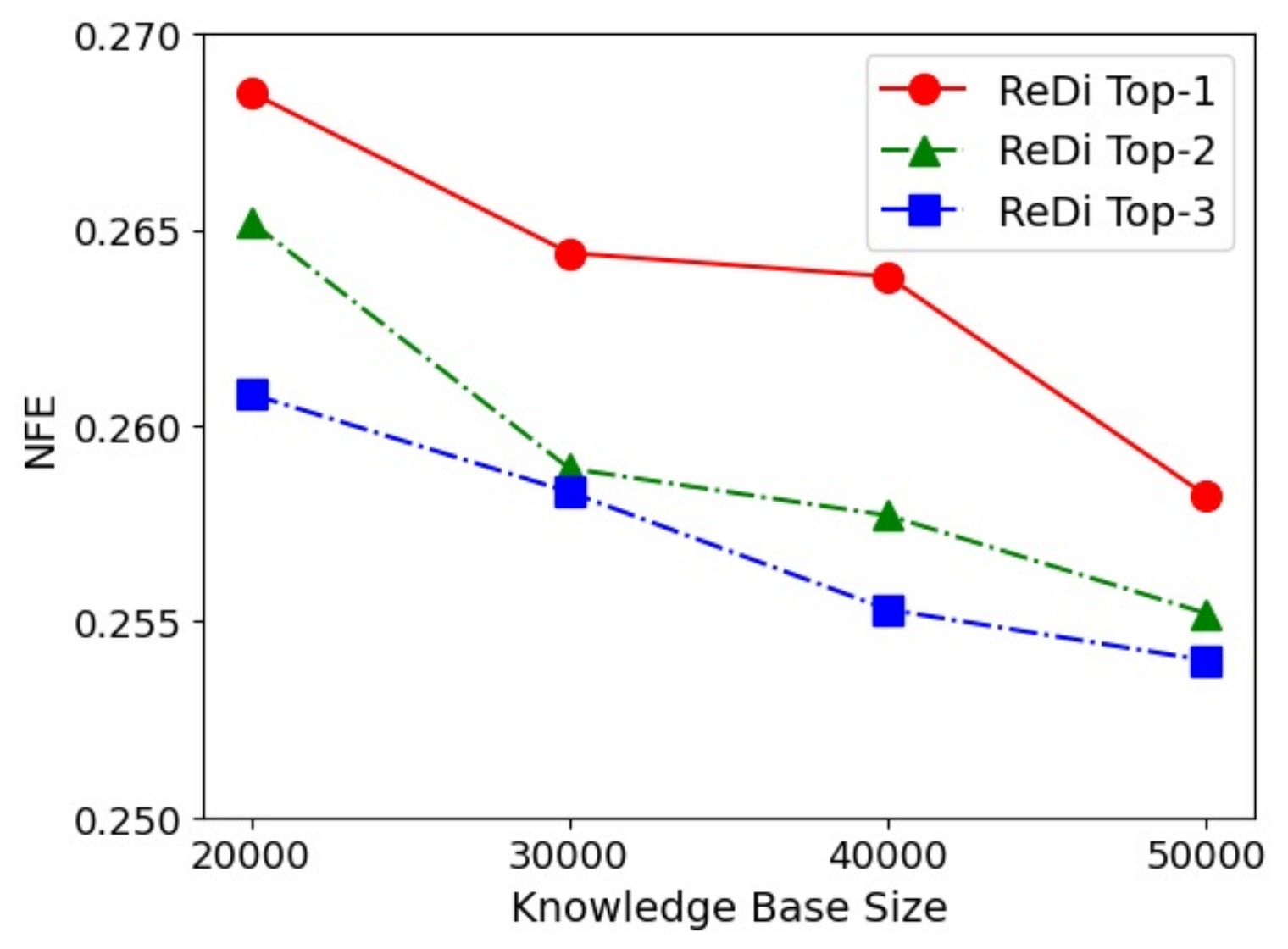}
         \caption{ReDi with different KB sizes}
         \label{fig:kbsize}
     \end{subfigure}
     \hfill
     \begin{subfigure}[b]{0.3\textwidth}
         \centering
         \includegraphics[width=\textwidth]{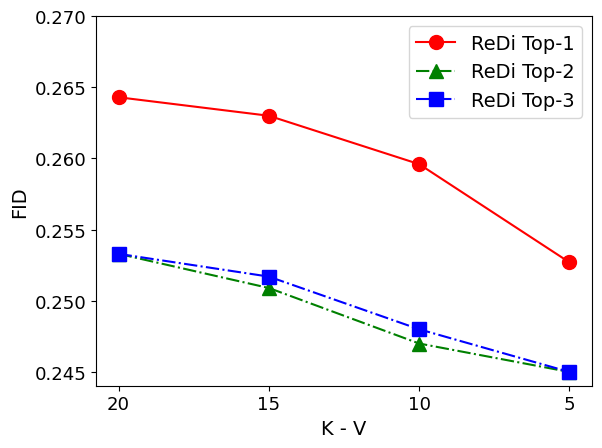}
         \caption{ReDi with different skip sizes}
         \label{fig:kv}
     \end{subfigure}
     \hfill
     \begin{subfigure}[b]{0.3\textwidth}
         \centering
         \includegraphics[width=\textwidth]{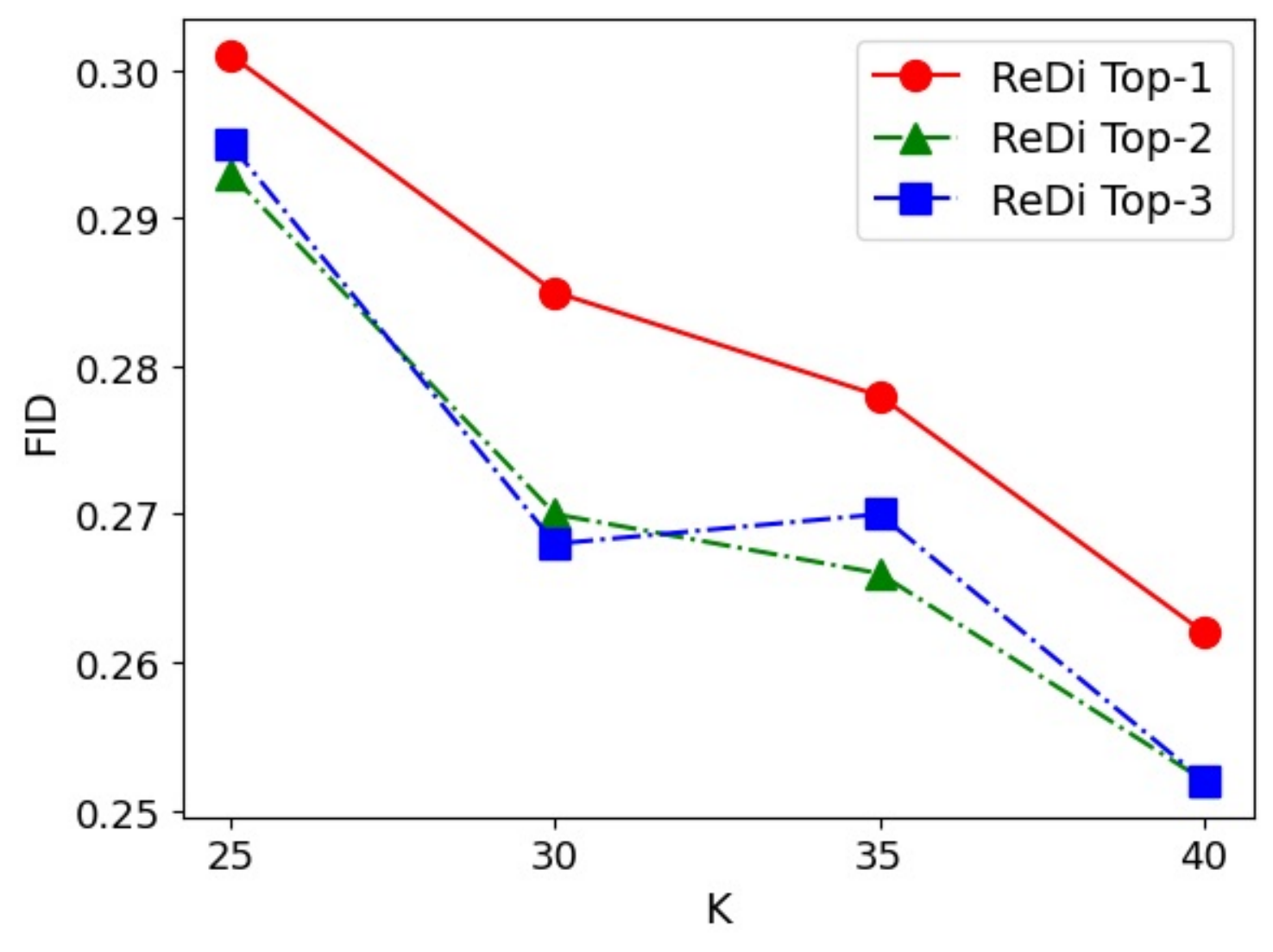}
         \caption{ReDi with key steps}
         \label{fig:kalone}
     \end{subfigure}
        \caption{The FID scores of ReDi applied to two numerical solvers for diffusion models, PNDM and DPM-Solver.}
        \label{fig:fid_abl}
\end{figure*}

\begin{new_content}
Furthermore, for stylistic prompts that can not be explicitly decomposed into $y^{\text{content}}$ and $y^{\text{style}}$, we propose to use prompt engineering techniques from the natural language processing community by asking a large language model to conduct the decomposition for us. We include an example of such method in Appendix \ref{sec:prompt}.
\end{new_content}

\section{The Tightness of the Theoretical Bound}

\label{sec:abl}

In this section, we conduct experiments of ReDi on PNDM to show that the proven bound is tight enough to be an estimator of the actual performance. The bound proven in \autoref{thm:main} is affected by the following factors:
\begin{itemize}[noitemsep, leftmargin=*, topsep=2pt]
    \item The distance to the retrieved neighbor $|\x_k-\hat\x_k|$ which depends on the size of the knowledge base $|\mathcal{B}|$ and the number of nearest neighbors $H$ used.
    \item The difference between the key step and the query step $k-v$, which depends on $k$ and $v$.
    \item The Lipschitz constant $L$ which depends on $k$ and $v$.
\end{itemize}

Therefore, we conduct ablation studies on the knowledge base size, the choice of $k$ and $v$, and the number of neighbors $H$ to check if the performance ReDi correlates well with the theoretical bound. In all ablation experiments, we use PNDM as the numerical sampler and evaluate the performance using FID scores on samples generated from MSCOCO validation. Our findings indicate that the performance of ReDi correlates well with the theoretical bound. Therefore the proven bound can be a good estimator for model performance.

\paragraph{Knowledge base size} We control the size of the knowledge base by randomly sampling from the complete knowledge base described in \autoref{sec:exp1}. We experiment with knowledge bases of sizes 20K, 30K, 40K and 50K. As shown in \autoref{fig:kbsize}, as the knowledge base size gets smaller, the performance of \method~drops. This finding is consistent with the theoretical bound since the bound is proportional to $|\x_k-\x_k'|$.

\paragraph{Key and query steps} There are two questions to be studied about how key and query steps influence the performance of ReDi - the difference between them $k-v$, and the choice of $k$. The bound from \autoref{thm:main} is proportional to the exponential of $k-v$. Therefore, we control the difference between the key and the query steps by fixing the key step $k=40$, and alternating $v$ to be $35, 30, 25$. As shown in \autoref{fig:kv}, as $K-V$ gets bigger, the performance of ReDi drops. This finding is consistent with the theoretical bound since the bound is proportional to $e^{k-v}$.
\noindent
The bound is also proportional to $e^L$. Even if the difference between $k$ and $v$ is fixed, the choice of $k$ alone can affect $L$ and the performance. We keep $k-v=15$ and alternate $k$ to be $40,30,20$. As shown in \autoref{fig:kalone}, as $k$ gets smaller, the performance of ReDi drops. This finding is consistent with the theoretical bound since the bound of Stable Diffusion explodes when $t\rightarrow0$ \cite{liu2021pseudo}.
\noindent
\paragraph{Number of neighbors retrieved} Increasing the number of neighbors can make the approximation of the query more accurate. Therefore, we control the number of neighbors in every ablation experiment to evaluate how $H$ affects the performance. As shown in \autoref{fig:fid_abl}, the performance of ReDi rises as the number of neighbors increases. This correlates well with the theoretical bound. We also find the difference between two $H$s converges as $H$ gets to $3$.

\begin{new_content}
\section{Discussion}

\paragraph{Generalizability with respect to guidance scale}
The knowledge base for \method~is built with the guidance scale of 7.5. We investigate whether \method~works for other guidance scales. The results and generated sample images are listed in Appendix \ref{sec:guidance_signal}. The results show that \method~is still able to function and produce similar-quality results when the guidance weight is not the same as the one used in the knowledge base.

\paragraph{Complicated and compositional prompts}

Due to the retrieval nature of \method, it is possible that when prompted with complicated and compositional prompts, there may not be a trajectory in $\mathcal{B}$ that's close enough to the prompt. To qualitatively evaluate \method's performance on complicated prompts, we pick the 4 longest prompts in the test set and compare the generated images of \method~and PNDM. The samples are shown in Appendix \ref{sec:long_prompts}. From the image samples with the longest prompts, we observe that ReDi does not show an inferior ability of compositionality compared with Stable Diffusion. Sometimes it shows better compositionality than Stable Diffusion. However, to enable better compositionality, it's better to have a larger knowledge base.

\paragraph{Impact on sample diversity} When the knowledge base is small, it's possible that the diversity of the generated samples can be limited to a smaller space around the data points in the knowledge base. We investigate how \method~affects the sample diversity by computing the Inception Score \citep{salimans2016improved} for the generated images and the ground truth images in Appendix \ref{sec:sample_diversity}. The results indicate that \method~is capable of generating diverse images.

\paragraph{Beyond the image modality} We have only utilized \method~under the text-to-image setting. However, it may be extended to other modalities. We argue that the extension is possible because DPM-solver, which has the same Lipschitz assumption as ours, is used in other domains. Our theoretical analysis (Theorem 5.3) is based on the Lipschitz assumption (Assumption 5.1), which is also the theoretical basis for the performance of the DPM-Solver \citep{lu2022dpm}. Variations of DPM-Solver have demonstrated effectiveness in diverse domains, including audio, video \citep{ruan2023mm}, and molecular graph \citep{huang2022conditional}. Therefore, we contend that \method~may also be applicable in these cases.
\end{new_content}

\section{Conclusion}

This paper proposes \textsc{ReDi}, a learning-free diffusion inference framework via trajectory retrieval. Unlike previous learning-free samplers that extensively explore more efficient numerical solvers, we focus on utilizing the sensitivity of the diffusion ODE, which leads to our choice of using partial trajectories as query and key for retrieval. We prove a bound for \textsc{ReDi} with the Lipschitz continuity of the diffusion model. The experiments on Stable Diffusion empirically verify \textsc{ReDi}'s capability to generate comparable images with improved efficiency. The zero-shot domain adaptation experiments shed light on further usage of ReDi and call for a better understanding of diffusion inference trajectories. We look forward to future studies built upon ReDi and the properties of the diffusion ODE. To make the best out of the limited knowledge base, it is also desirable to study ReDi in a compositionality setting so that the combination of different trajectories can be more controllable.

\begin{new_content}

\section*{Acknowledgements}

This work is partially supported by unrestricted gifts from IGSB and Meta via the Institute for Energy Efficiency.

We thank Weixi Feng, Tsu-Jui Fu, Jiabao Ji, Yujian Liu and Qiucheng Wu for helpful discussions and proof-reading.

\end{new_content}

\bibliography{example_paper}
\bibliographystyle{icml2023}

 \newpage
 \appendix
 \onecolumn
\section{Appendix}

\subsection{Proof of ODE Sensitivity} 
\label{sec:a_proof}

\begin{theorem} \label{thm:gron}
(Grönwall–Bellman inequality \cite{gronwall1919note, bellman1943stability}) Suppose $\lambda: [a, b] \to \mathbb{R}$ be continuous and $\mu: [a, b] \to \mathbb{R}$ be continuous and non-negative. Let a continuous function $y: [a, b] \to \mathbb{R}$ and for $t \in [a, b]$ it satisfies
$$ y(t) \le \lambda(t) + \int_{a}^{t} \mu(s) y(s) \,\d s ,$$
then for t on the same interval
$$ y(t) \le \lambda(t) + \int_{0}^{t} \lambda(s) \mu(s) e^{\int_{s}^{t} \mu(\tau) \,d\tau } \,\d s $$
\end{theorem}

\begin{theorem} \label{thm:sensitivity}
(Sensitivity of ODE \cite{khalil2002nonlinear}) 

Let $f(t, x)$ be piecewise continuous in t and L-Lipschitz in $x$ on $[t_0, T] \times R^n $, and let $y: \mathbb{R} \to \mathbb{R}^n$ is the solution of 
$$ \dot{y} = f(t, y), \quad y(t_0) = y_0 $$
and $z: \mathbb{R} \to \mathbb{R}^n$ is the solution of 
$$ \dot{z} = f(t, z) + g(t, z), \quad z(t_0) = z_0 $$
with respect to $y(t), z(t) \in \mathbb{R}^n $ for all $t \in [t_0, T]$. Suppose there exists $\mu > 0$ such that 
$|g(t, x)| \leq \mu $ for all $(t, x) \in [t_0, T] \times \mathbb{R}^n $
and further suppose $ | y_0 - z_0 | = \gamma $. Then for any $t \in [t_0, T]$,
$$ \mid y(t) - z(t) \mid \leq \gamma e^{L(t-t_0)} + \frac{\mu}{L} (e^{L(t-t_0)} - 1).$$

\textit{Proof: } First we can write the solutions of y and z by
$$ y(t) = y_0 + \int_{t_0}^{t} f(s, y(s))\,\d s,$$
$$ z(t) = z_0 + \int_{t_0}^{t} (f(s, z(s))) + g(s, z(s))\,\d s.$$
Subtracting the above two solutions yields
$$ \mid y(t) - z(t) \mid \leq \mid y_0 - z_0 \mid + \int_{t_0}^{t} \mid g(s, z(s)) \mid \,\d s + \int_{t_0}^{t} \mid f(s, y(s)) - f(s, z(s) \mid \,\d s 
\leq \gamma + \mu (t - t_0) + \int_{t_0}^{t} L \mid y(s) - z(s) \mid \,\d s.$$
By letting $\lambda (t) = \gamma + \mu (t - t_0)$ and $\mu (s) = L$, then applying the \ref{thm:gron} to the function $|y(t) - z(t)|$ gives,
$$ \mid y(t) - z(t) \mid \leq \gamma + \mu (t - t_0) + \int_{t_0}^{t} L(\gamma + \mu (s - t_0)) e^{L(t - s)} \,\d s.$$
Finally, integrating the last term on the right hand gives,
$$ \mid y(t) - z(t) \mid \leq \gamma + \mu (t - t_0) - \gamma - \mu (t - t_0) + \gamma e^{L(t - t_0)} 
+ \int_{t_0}^{t} \mu e^{L(t - s)} \,\d s $$
$$ = \gamma e^{L(t - t_0)} + \frac{\mu}{L}(e^{L(t - t_0)} - 1),$$
which finishes the proof. 
\end{theorem}
In our setting, the $g(t, z)$ is simply equal to $0$, so we have the following inequality
$$ \mid y(t) - z(t) \mid \leq \gamma e^{L(t - t_0)}.$$

\begin{new_content}
\newpage
\subsection{\method's Performance with Different Guidance Signals}
\label{sec:guidance_signal}
We list the FID scores of \method~with different guidance scales in \autoref{tab:hoho}. We also show generate samples using different scales in \autoref{fig:guidance_samples}.
\begin{table}[h!]
\begin{new_content}
\caption{\upd{The FID scores $\downarrow$ of \textsc{ReDi} with different guidance scales. Note that the knowledge base is built with a guidance scale of 7.5. the performance of ReDi experiences a slight drop when the guidance weight differs from the one used in the knowledge base.}}
\label{tab:hoho}
\vskip 0.15in
\begin{center}
\begin{sc}
\begin{tabular}{llllll}
    \hline
Guidance    & 5.5                              & 6.5                              & 7.5        & 8.5 & 9.5                      \\ \hline FID & 0.271 & 0.269 & 0.264 & 0.268 & 0.267   \\ \hline
\end{tabular}
\end{sc}
\end{center}
\end{new_content}
\end{table}

\begin{figure*}[h!]
     \centering
     \begin{subfigure}[b]{0.47\textwidth}
         \centering
         \includegraphics[width=\textwidth]{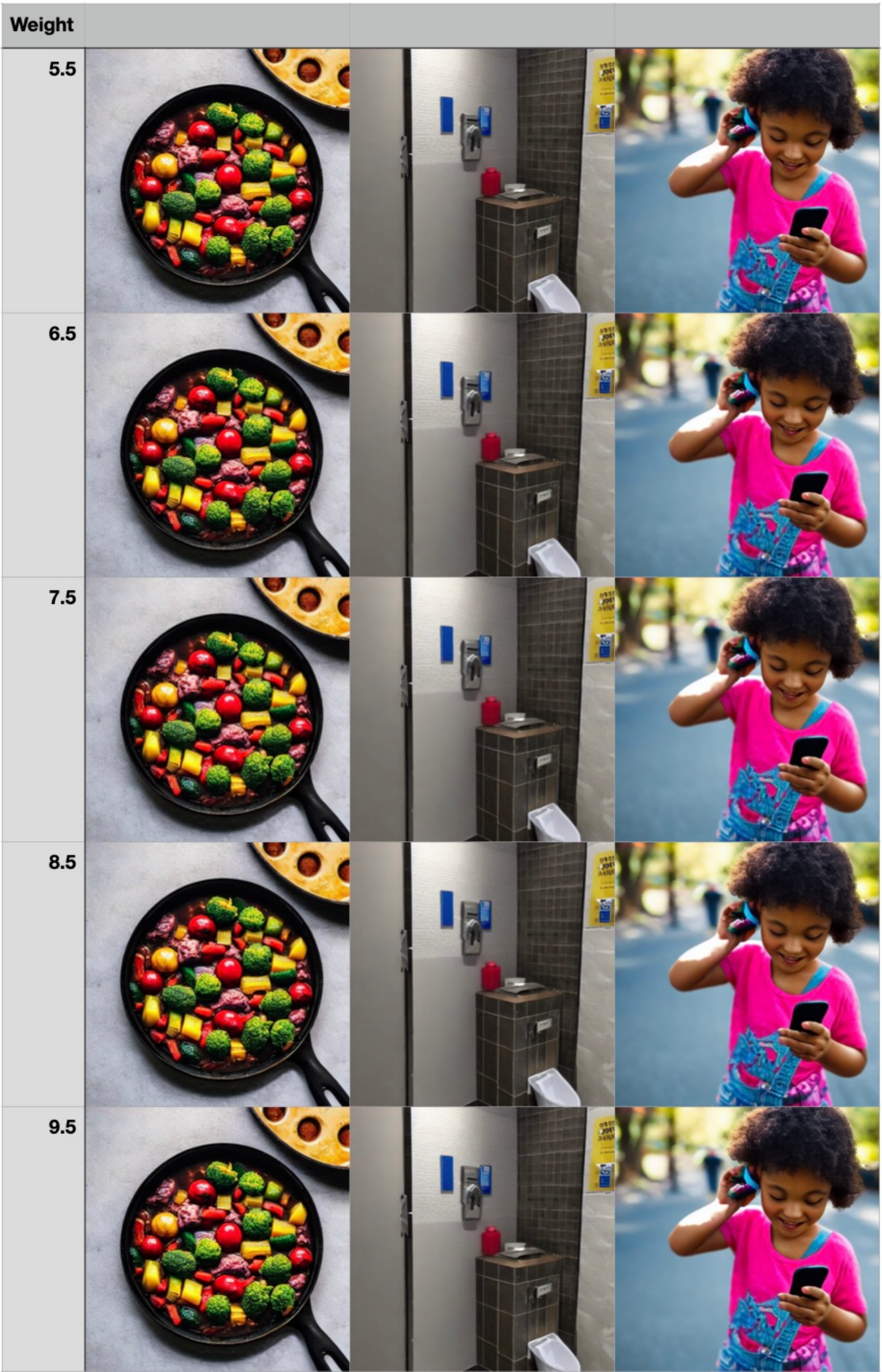}
     \end{subfigure}
     \hfill
     \begin{subfigure}[b]{0.47\textwidth}
         \centering
         \includegraphics[width=\textwidth]{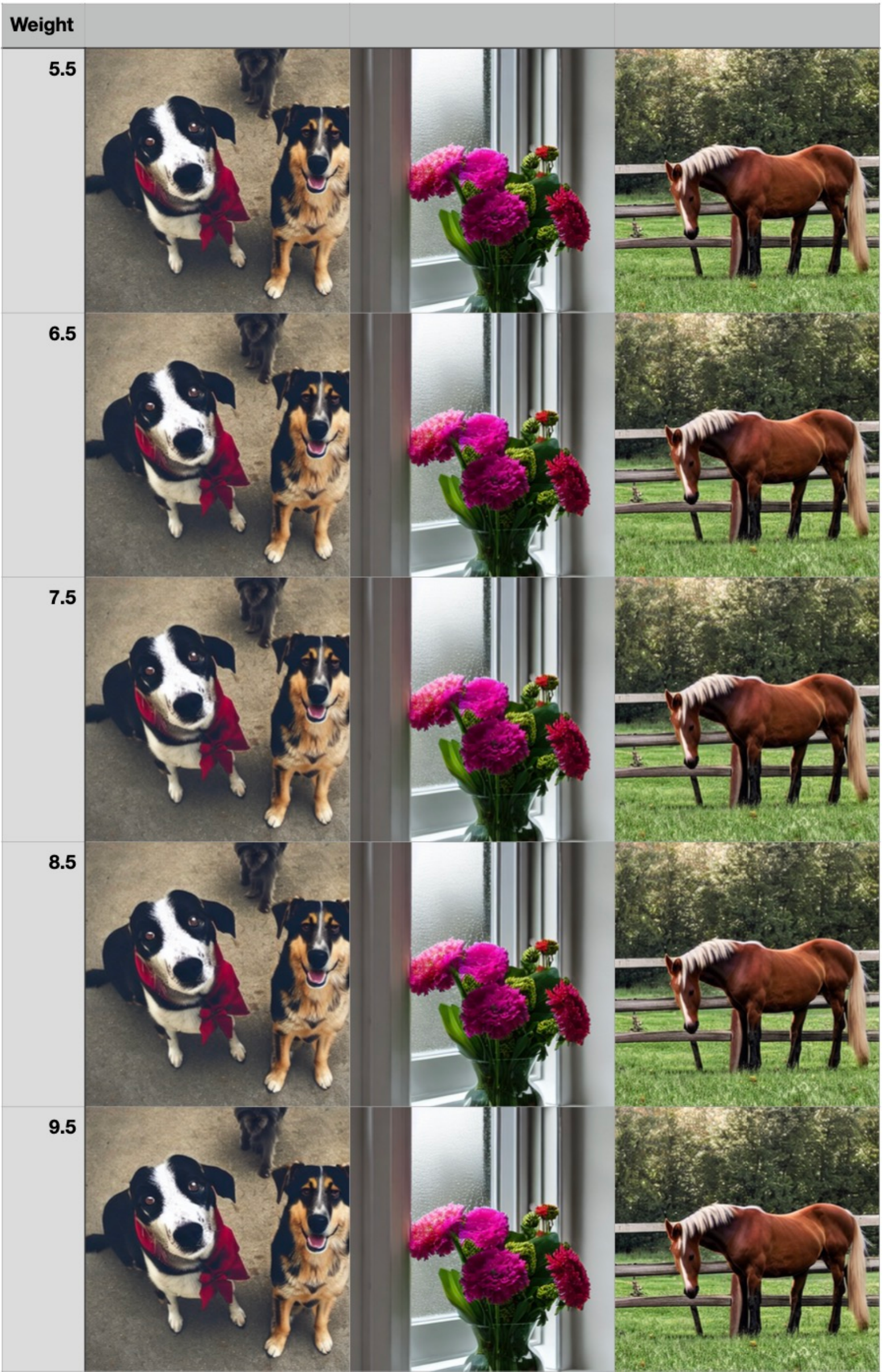}
     \end{subfigure}
     \caption{\method-generated images with different guidance scales.}
     \label{fig:guidance_samples}
\end{figure*}

\newpage
\subsection{Samples with Complicated Prompts Generated by \method}
\label{sec:long_prompts}
We show the image samples generated by \method~and PNDM from the longest prompts in \autoref{fig:long_prompts}. ReDi does not show an inferior ability of compositionality compared with Stable Diffusion. Sometimes it shows better compositionality than Stable Diffusion. For example, In row 3, Stable Diffusion incorrectly generated a mask covering the rider's face when it should be covering the horse's face. In row 4, Stable Diffusion failed to generate the catcher and umpire in the background, whereas ReDi successfully did. These findings demonstrate the potential of ReDi to handle complex prompts.
\begin{figure*}[h!]
     \centering
         \includegraphics[width=0.65\textwidth]{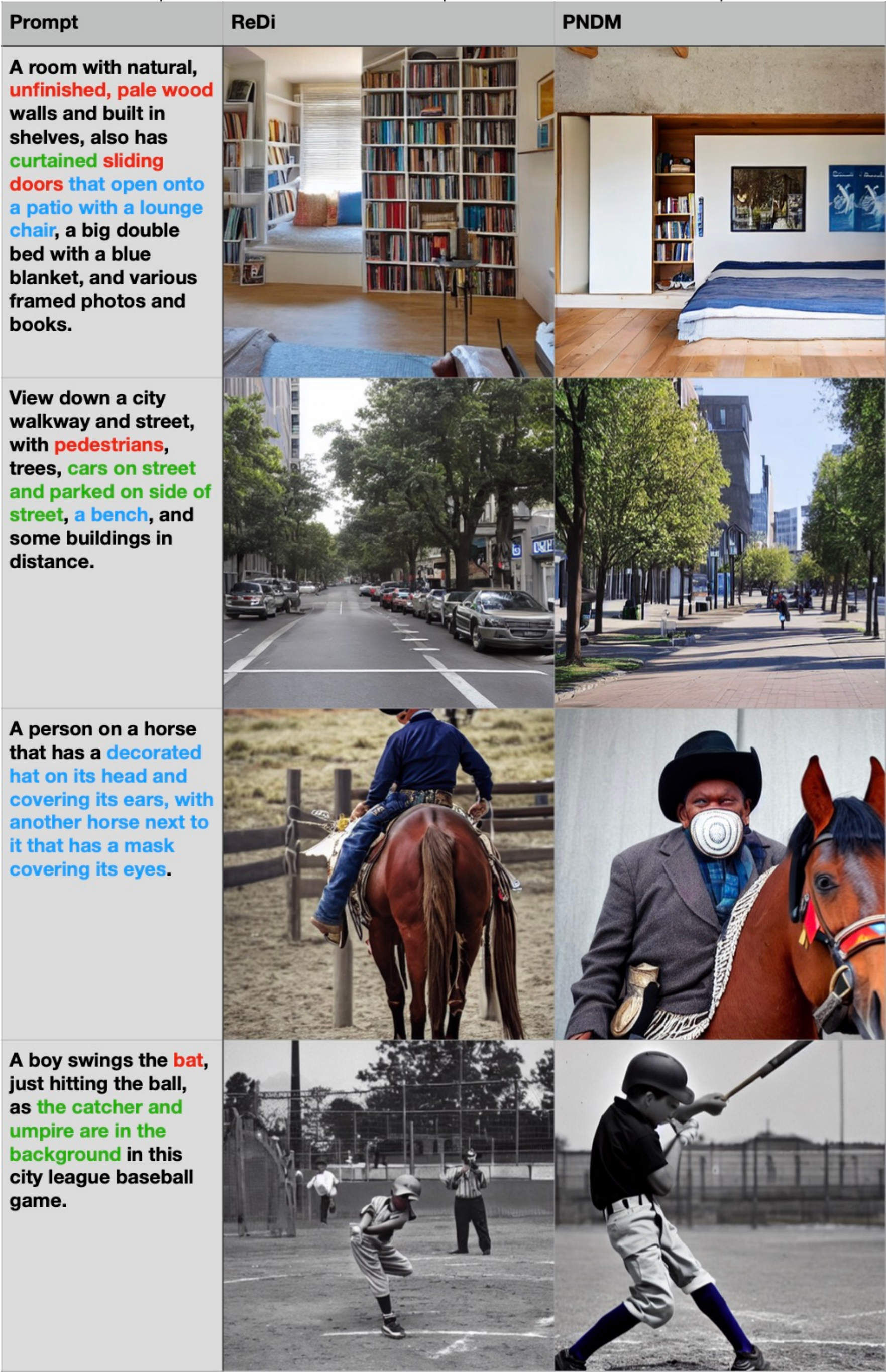}
     \caption{\method-generated images with the longest prompts. \textcolor{red}{Red} stands for the parts \method~missed. \textcolor{green}{Green} stands for the part PNDM missed. \textcolor{blue}{Blue} stands for the parts both methods missed.}
     \label{fig:long_prompts}
\end{figure*}

\newpage
\subsection{Evaluation of Sample Diversity}
\label{sec:sample_diversity}
As shown in \autoref{tab:sample_diversity}, the Inception Scores for \method~are comparable to those of the ground truth images in the MS-COCO validation set. Based on these results, we argue that \method~is capable of generating diverse images.

\begin{table}[h!]
\begin{new_content}
\caption{\upd{The Inception Scores $\uparrow$ of \textsc{ReDi} with different NFEs compared to the Inception Score of the ground truth images. The Inception Score is maximized when the images are evenly distributed and diverse.}}
\label{tab:sample_diversity}
\vskip 0.1in
\begin{center}
\begin{sc}
\begin{tabular}{llll}
    \hline
NFE & 20 & 30 & 40
                      \\ \hline \method & 29.49 & 30.79 & 31.23  \\ \hline Ground Truth & 30.79 & 30.79 & 30.79 \\ 
                      \hline 
\end{tabular}
\end{sc}
\end{center}
\end{new_content}
\end{table}
\subsection{The Initial Sample in Knowledge Base Construction}
\label{sec:initial_sample}

In Algorithm \autoref{alg:conskb}, the sample $\x_T$ to initialize every trajectory in the knowledge base is sampled from a conditional distribution $p(\x_T|\x^{(i)}$. However, the noise schedule in diffusion models causes the signal-to-noise ratio to approach $0$ at time step $T$. This makes sampling from $p(\x_T|\x^{(i)})$ approximately the same as sampling from the unconditional distribution $p(\x_T)$. This makes it possible for \method~to construct the knowledge base without the ground truth images and with only the prompts. We investigate this possibility by rebuilding the knowledge base with initial samples from the unconditional distribution. The results reported in \autoref{tab:initial_sample} indicate that whether the initial distribution is conditional does not significantly affect the performance of \method.

\begin{table}[h!]
\begin{new_content}
\caption{\upd{The FID Scores $\downarrow$ of \textsc{ReDi} when the knowledge base is built with the unconditional initial distribution and the conditional one.}}
\label{tab:initial_sample}
\vskip 0.1in
\begin{center}
\begin{sc}
\begin{tabular}{llll}
    \hline
NFE & 20 & 30 & 40
                      \\ \hline $\x_T\sim p(\x_T)$ & 0.268 & 0.263 & 0.265  \\ \hline $\x_T \sim p(\x_T|\x_0)$ & 0.265 & 0.264 & 0.262 \\ 
                      \hline 
\end{tabular}
\end{sc}
\end{center}
\end{new_content}
\end{table}
\subsection{Validation of Effectiveness}
\label{sec:effectiveness}

\paragraph{Vanilla skipping from $k$ to $v$} We compare it \method~with a naive substitute - directly approximation of $\x_v$ from $\x_k$ using any numerical solver, without resorting to retrieval. The average L2 distances to the true value $\x_v$ for both the \method~retrieved $\hat \x_v$ and the vanilla skipped $\tilde x_v$ is reported in \autoref{tab:vanilla_skip}. It can be noted that \method~retrieved values are much closer to the true value, indicating the effectiveness of \method. We also compute the FID for vanilla skipping under the PNDM $k=40, v=20$ setting. The FID of vanilla skipping is 1.58, which is much worse than the 0.267 from \method.

\begin{table}[h!]
\begin{new_content}
\caption{\upd{The L2 distances between the true value $\x_v$ and the estimated values.}}
\label{tab:vanilla_skip}
\vskip 0.1in
\begin{center}
\begin{sc}
\begin{tabular}{llll}
    \hline
NFE & 20 & 30 & 40
                      \\ \hline $d(\x_v, \hat \x_v)$ & 7.95 & 18.73 & 7.95  \\ \hline $d(\x_v,\tilde \x_v)$ & 44.20 & 24.16 & 44.20 \\ 
                      \hline 
\end{tabular}
\end{sc}
\end{center}
\end{new_content}
\end{table}

\paragraph{Similarity with Original Samples} We compute the distance between \method-generated samples and samples generated by the original sampler and list them in \autoref{tab:fid_dist}. The distance between ReDi and the original samples is very small (with an average ~25\% percentage) compared to the distance between the original samples and the ground truth. Therefore we argue that ReDi does not introduce much error to the original sampling trajectory.

\begin{table}[t!]
\begin{new_content}
\caption{\upd{The FID between the \method~and original samples, original samples and ground truth.}}
\label{tab:fid_dist}
\vskip 0.1in
\begin{center}
\begin{sc}
\begin{tabular}{llll}
    \hline
NFE & 20 & 30 & 40
                      \\ \hline FID(\method, Original) & 0.0745 & 0.0655 & 0.0632  \\ \hline FID(Original, Ground Truth) & 0.274 & 0.268 & 0.272 \\ 
                      \hline Percentage & 27.0\% & 24.4\% &  24.1\% \\ \hline
\end{tabular}
\end{sc}
\end{center}
\end{new_content}
\end{table}
\newpage
\subsection{Prompt Engineering for Explicit Style-Content Decomposition}
\label{sec:prompt}

We demonstrate how we can automatically decompose an out-of-domain prompt into $y^{\text{in}}$ and $y^{\text{out}}$ with the prompt \textit{``A photo of a pig wearing an astronaut hat.''}.

Unlike prompts with stylistic suffixes, this prompt cannot be easily split into an in-domain prefix and an out-of-domain suffix. Even with the original diffusion model, the generation quality is not satisfactory, as shown in \autoref{fig:sd_pig}.

\begin{figure*}[ht]
\begin{centering}
     \begin{subfigure}[b]{0.3\textwidth}
         \centering
         \includegraphics[width=\textwidth]{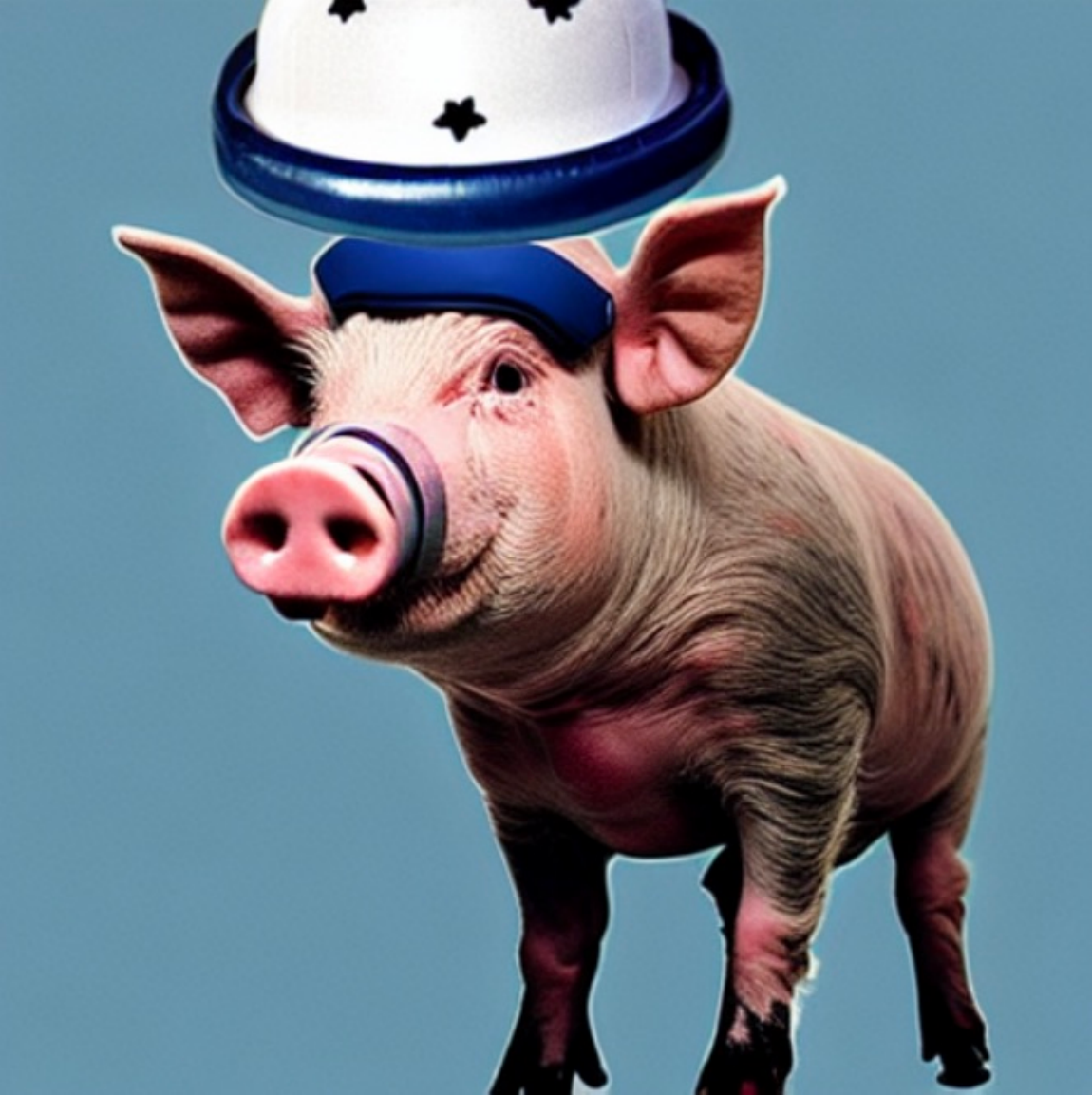}
         \caption{Sample from Stable Diffusion}
         \label{fig:sd_pig}
     \end{subfigure}
     \begin{subfigure}[b]{0.3\textwidth}
         \centering
         \includegraphics[width=\textwidth]{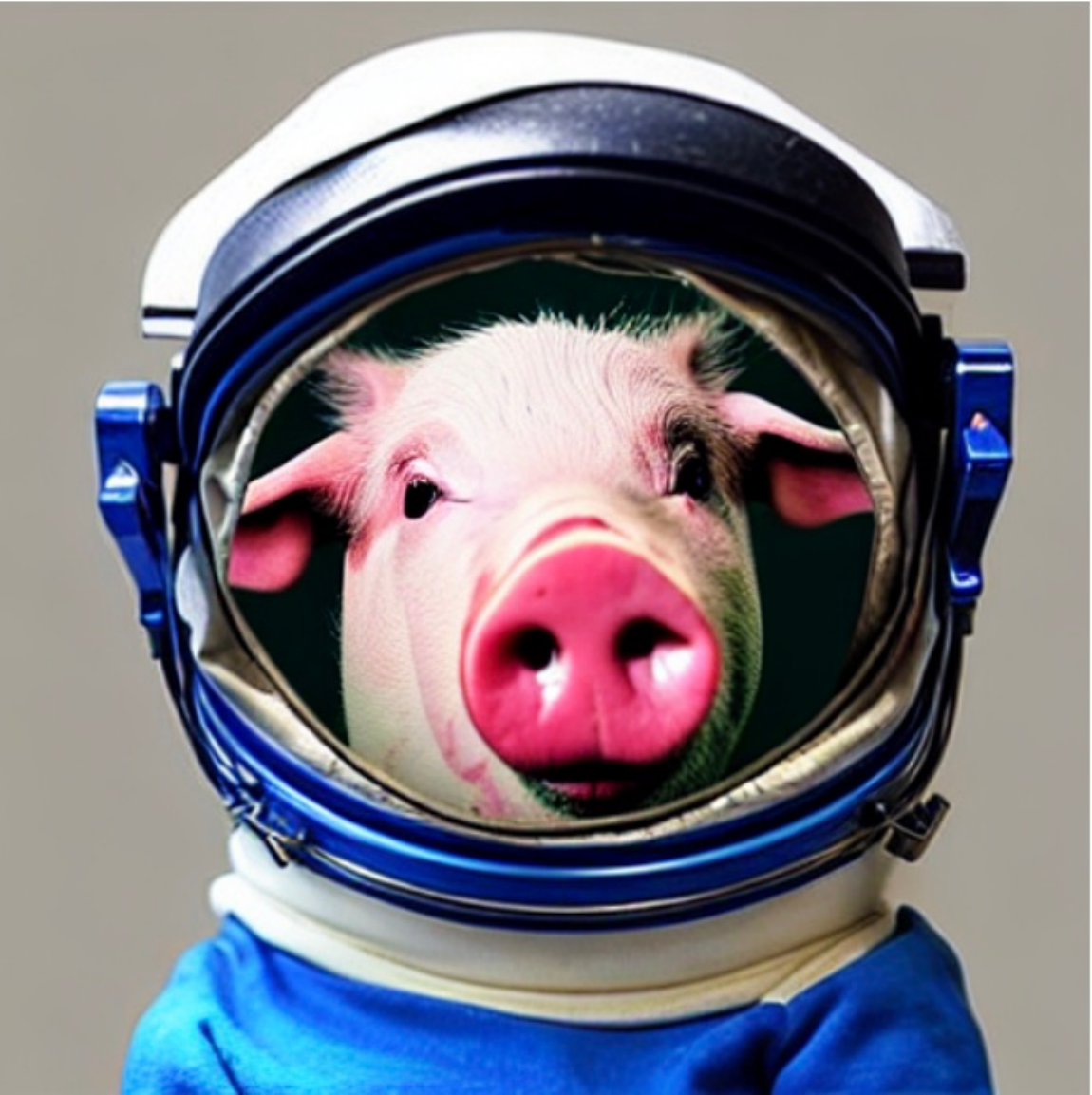}
         \caption{Sample from ReDi}
         \label{fig:redi_pig}
     \end{subfigure}
        \caption{Generation samples from Stable Diffusion and ReDi using the prompt \textit{``A photo of a pig wearing an astronaut hat''}.}
\end{centering}

    \label{fig:gpt4_samples}
\end{figure*}

We then asked GPT-4 to re-write the prompt with minimal changes to a more usual one (which is used as $y^{\text{in}}$) and then generate the image
with ReDi. The response from GPT-4 is shown in \autoref{fig:gpt-4}. GPT-4 is able to find out what is unusual about the prompt and re-write it to ``a photo of a person wearing an astronaut hat''. With that, \method~can generate much better samples as shown in \autoref{fig:redi_pig}.

\begin{figure*}[h]
\begin{centering}
         \includegraphics[width=0.5\textwidth]{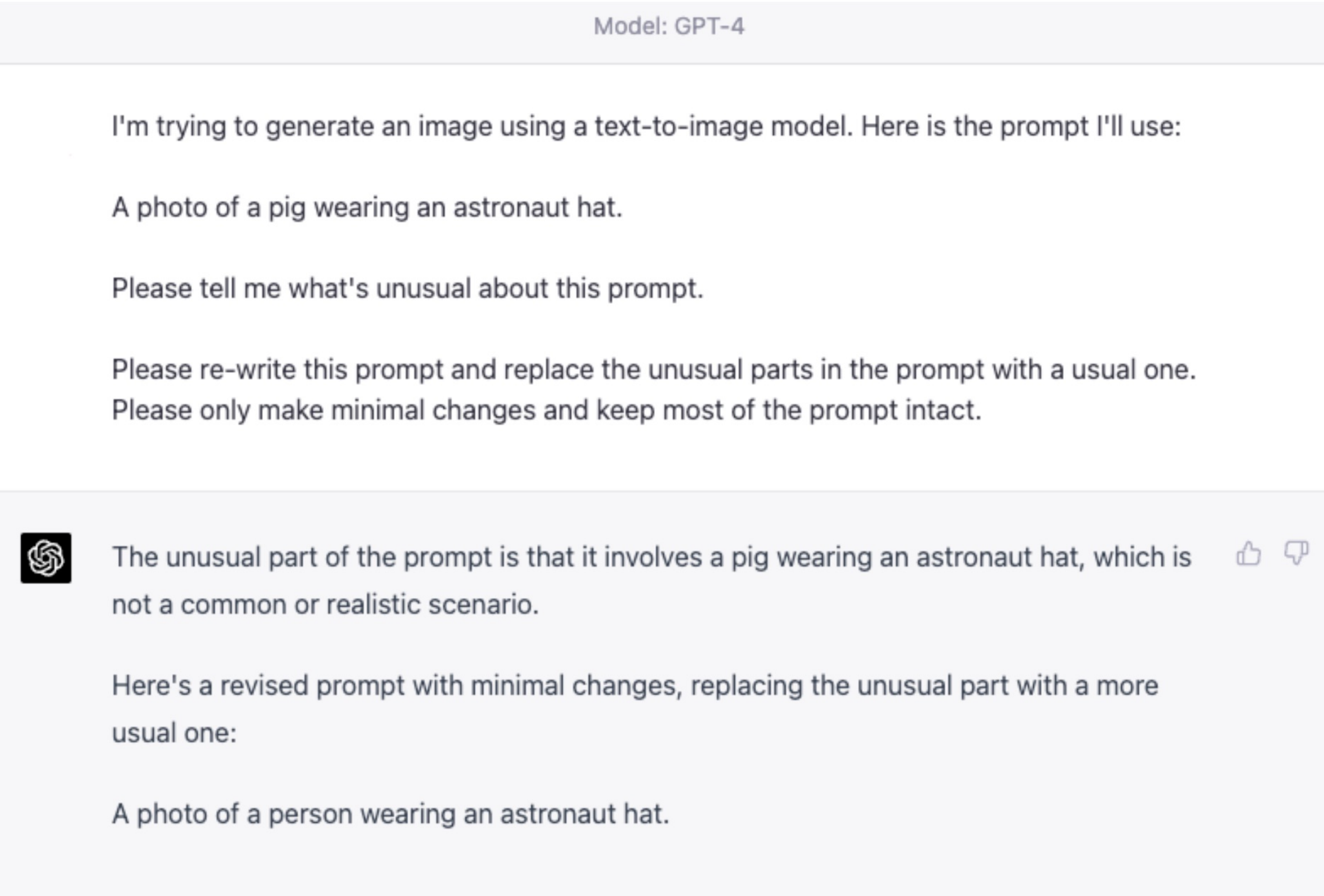}
         \caption{Response from GPT-4.}
         \label{fig:gpt-4}
\end{centering}
\end{figure*}
\end{new_content}

\end{document}